\documentclass[journal]{IEEEtran}

\usepackage[
    bookmarks=true,
    hypertexnames=false
]{hyperref}

    \usepackage{cite}
    \usepackage{amsmath}
    \usepackage{amsthm}
    \usepackage{amsfonts}
    \usepackage{amssymb}
    \usepackage{graphicx}
    \usepackage{array}
    \usepackage[dvipsnames]{xcolor}
    \usepackage{subcaption}
    \usepackage{mathrsfs}
    \usepackage{multicol}
    \usepackage{multirow}
    \usepackage{arydshln}
    \usepackage{tabularx}
    \usepackage{calc}

        \newcommand{\abs}[1]{\left|#1\right|}

        \newcommand{\card}[1]{\left|#1\right|}
        \newcommand{\set}[1]{\left\{#1\right\}}

        \newcommand{\reals}{\mathbb{R}}
        \newcommand{\R}{\reals}

        \newcommand{\norm}[1]{\abs{\abs{#1}}}
        \newcommand{\tpose}{^{T}}

        \newcommand{\mc}[1]{\mathcal{#1}}
        \newcommand{\ms}[1]{\mathscr{#1}}
        \newcommand{\mb}[1]{\mathbb{#1}}

    \usepackage{letltxmacro}
    \LetLtxMacro\orgvdots\vdots
    \LetLtxMacro\orgddots\ddots

    \makeatletter
    \DeclareRobustCommand\vdots{%
        \mathpalette\@vdots{}%
    }
    \newcommand*{\@vdots}[2]{%
        \sbox0{$#1\cdotp\cdotp\cdotp\m@th$}%
        \sbox2{$#1.\m@th$}%
        \vbox{%
            \dimen@=\wd0 %
            \advance\dimen@ -3\ht2 %
            \kern.5\dimen@
            \dimen@=\wd2 %
            \advance\dimen@ -\ht2 %
            \dimen2=\wd0 %
            \advance\dimen2 -\dimen@
            \vbox to \dimen2{%
                \offinterlineskip
                \copy2 \vfill\copy2 \vfill\copy2 %
            }%
        }%
    }
    \DeclareRobustCommand\ddots{%
        \mathinner{%
            \mathpalette\@ddots{}%
            \mkern\thinmuskip
        }%
    }
    \newcommand*{\@ddots}[2]{%
        \sbox0{$#1\cdotp\cdotp\cdotp\m@th$}%
        \sbox2{$#1.\m@th$}%
        \vbox{%
            \dimen@=\wd0 %
            \advance\dimen@ -3\ht2 %
            \kern.5\dimen@
            \dimen@=\wd2 %
            \advance\dimen@ -\ht2 %
            \dimen2=\wd0 %
            \advance\dimen2 -\dimen@
            \vbox to \dimen2{%
                \offinterlineskip
                \hbox{$#1\mathpunct{.}\m@th$}%
                \vfill
                \hbox{$#1\mathpunct{\kern\wd2}\mathpunct{.}\m@th$}%
                \vfill
                \hbox{$#1\mathpunct{\kern\wd2}\mathpunct{\kern\wd2}\mathpunct{.}\m@th$}%
            }%
        }%
    }
    \makeatother

\usepackage{cleveref}
\Crefname{figure}{Fig.}{Figs.}
\Crefname{equation}{Eq.}{Eqs.}
\Crefname{lemma}{Lemma}{Lemmata}
\Crefname{proposition}{Proposition}{Propositions}
\Crefname{assumption}{Assumption}{Assumptions}
\Crefname{theorem}{Theorem}{Theorems}
\Crefname{section}{Section}{Sections}
\Crefname{subsection}{Subsection}{Subsections}
\Crefname{appendix}{Appendix}{Appendices}
\Crefname{corollary}{Corollary}{Corollaries}

\DeclareMathOperator{\SO}{SO}
\DeclareMathOperator{\SE}{SE}
\newcommand{\ST}{\mathrm{s.t.}}
\DeclareMathOperator{\des}{des}
\newcommand{\colsep}{\hspace{0.5em}}
\DeclareMathOperator{\lb}{lb}
\DeclareMathOperator{\ub}{ub}
\DeclareMathOperator{\IK}{IK}
\newcommand{\monogram}[3]{{}^{#2}\!#1^{#3}}
\DeclareMathOperator{\SDF}{SDF}
\newcommand{\tableNA}{\textcolor{gray}{N/A}}
\DeclareMathOperator{\com}{com}
\DeclareMathOperator{\Conv}{Conv}
\DeclareMathOperator{\find}{find}

\newcolumntype{C}{>{\centering\arraybackslash}X}
\newcolumntype{V}{!{\vrule width 1.2pt}}
\newcommand{\goldstar}{\textcolor{BurntOrange}{$\bigstar$}}

\usepackage{color-edits}

\addauthor{tc}{cyan}
\addauthor{lt}{orange}
\definecolor{darkgreen}{rgb}{0.0, 0.75, 0.0}
\addauthor{aa}{darkgreen}
\addauthor{rt}{red}

\addauthor{hd}{purple}
\addauthor{ss}{teal}

\begin{document}

\title{A Framework for Combining Optimization-Based and Analytic Inverse Kinematics}

\author{Thomas Cohn$^\star$, Lihan Tang$^\star$, Alexandre Amice, and Russ Tedrake%
\thanks{$^\star$ denotes equal contribution. The authors are with the Computer Science and Artificial Intelligence Laboratory (CSAIL), Massachusetts Institute of Technology, 32 Vassar St, Cambridge, MA, 02139. \texttt{[tcohn,tangles,amice,russt]@mit.edu}}}

\maketitle

\begin{abstract}
Analytic and optimization methods for solving inverse kinematics (IK) problems have been deeply studied throughout the history of robotics.
The two strategies have complementary strengths and weaknesses, but developing a unified approach to take advantage of both methods has proved challenging.
A key challenge faced by optimization approaches is the complicated nonlinear relationship between the joint angles and the end-effector pose.
When this must be handled concurrently with additional nonconvex constraints like collision avoidance, optimization IK algorithms may suffer high failure rates.
We present a new formulation for optimization IK that uses an analytic IK solution as a change of variables, and is fundamentally easier for optimizers to solve.
We test our methodology on three popular solvers, representing three different paradigms for constrained nonlinear optimization.
Extensive experimental comparisons demonstrate that our new formulation achieves higher success rates than the old formulation and baseline methods across various challenging IK problems, including collision avoidance, grasp selection, and humanoid stability.

\end{abstract}


\section{Introduction}
\label{sec:introduction}
\IEEEPARstart{I}{nverse} kinematics (IK) is a fundamental problem in robotics, and has been extensively studied throughout the entire history of the field~\cite{pieper1969kinematics,herrera1988symbolic,raghavan1990kinematic,diankov2010ikfast,lakshmi2026decade}.
For a kinematic chain of rigid links connected by movable joints, IK seeks to compute the positions of the individual joints so as to achieve a desired Cartesian pose for the end-effector of the chain.
IK is a fundamental primitive in robotic manipulation and used in myriad domains, from warehouses and factory floors to human-centric environments to homes and stores.
IK is also frequently used in computer graphics, for art, video games, and animation~\cite{aristidou2018inverse}.
Finally, IK has even seen use in the fields of rehabilitative medicine~\cite{qamar2014adding,bodo2022comparative}, and for the purposes of studying human motion~\cite{pizzolato2017real,liu2020human}.

Early study of IK largely focused on symbolic analysis.
The pose of the end-effector of a kinematic chain can naturally be written down as a function of its joint angles.
Determining joint angles that achieve a desired end-effector pose amounts to solving a highly-structured system of trigonometric polynomial equations.
In certain special cases, this system can be directly solved via geometric or symbolic analysis~\cite{pieper1969kinematics}.
When the number of degrees of freedom matches the dimension of the constraint (e.g., when fixing the pose of the end-effector of a robot arm with 6 revolute joints in 3D space), there are finitely many solutions in general.
But with extra degrees of freedom, the system becomes underdetermined, leading to an infinitely many solutions.
This can be handled via joint-locking, sampling, or by taking in additional \emph{self-motion} parameters to disambiguate the solutions~\cite{gottlieb1988topology,luck1993self,diankov2010ikfast,faria2018iiwa,he2021frankapanda}.

With the continued improvements to numerical computing over the past decades, numerical algorithms for solving IK have become incredibly prevalent~\cite{buss2004introduction}.
Such approaches are appealing for several reasons.
They are general, as opposed to an analytic solution, which are specific to certain classes of robot arms.
Numerical approaches are also much more capable of handling kinematic redundancy, and they can be used to solve a broader variety of inverse kinematics problems beyond the case of fixing the end-effector pose.
For example, we might allow the target pose to vary slightly, or ask for a least-squares solution when the actual goal is unreachable.
With modern numerical computing, these algorithms can easily be used in realtime applications.

As these numerical IK algorithms have improved and become more general, the problem itself has broadened in scope.
The IK problem can be embedded within other problems, such as grasping, motion planning, and even complex goals like stability for humanoid robots~\cite{dai2014whole,fallon2015architecture,marion2018director}.
In its most general form, IK can be seen as a broad class of kinematic optimization problems~\cite[\S 6.1]{russtedrake2024manipulation}, and the wealth of papers promoting this perspective speak to its continued success.

However, optimization IK is not without its challenges.
Imposing the IK constraint means the optimizer must find a solution satisfying a complicated nonlinear equality constraint.
When adding additional constraints to the problem, such as collision avoidance or humanoid stability, optimizers may get stuck in local minima, or even fail to return a feasible guess.
Collision avoidance is known to be particularly challenging, due to its inherent nonconvexity and many infeasible local minima.

In summary, the current state of the art can largely be separated into two distinct lines of work: analytic and numerical (optimization) approaches.
Currently, these approaches are largely incompatible, but each has different advantages.
Analytic IK is incredibly fast and reliable, and optimization IK is flexible and more general.
Our proposed, combined method is motivated by the desire for the positives of both approaches.

\subsection{Our Contributions}

We present a unifying approach for solving optimization-style IK problems for systems that have an analytic solution.
We treat the analytic IK function as a smooth change of variables, so that the decision variables in the optimization problem are the pose of the end-effector and the self-motion parameters.
The joint angles can be written as functions of these variables with the analytic IK mapping, allowing us to impose the original costs and constraints.
In particular, solving such optimization problems with gradient-based solvers is enabled by automatic differentiation of the analytic IK function, a strategy presented in~\cite{cohn2024constrained}.

The proposed formulation trades off simplicity of the IK constraint for complexity in other costs and constraints.
For example, simple costs like joint centering and constraints like joint limits become complicated, nonlinear expressions in the new coordinate system.
Additionally, further domain constraints need to be imposed in the new coordinates to ensure that our variables remain within the domain of the analytic IK mapping.
On the other hand, the IK constraint becomes \emph{linear} in the new decision variables. 

The relative benefits of this trade-off for optimization-based formulations becomes apparent when inspecting the assumptions of the majority of standard nonlinear solvers.
The majority of algorithms operate either in the entire ambient space, or in the relative interior of the problem domain, i.e. in the intersection of the linear equality constraints with the interior of the inequality constraints.
The former either precludes the use of nonlinear equality constraints altogether as is the case for AGS~\cite[\S 8.1]{strongin2013ags}, CMA-ES~\cite[\S B.5]{hansen2016cma}, and NFQPIM~\cite{shang2018nfqpim} or requires converting the equality to a pair of inequalities, thus resulting in a failure of the Linear Independence Constraint Qualification (LICQ) that many solvers rely on for numerical stability~\cite[\S 4]{gill2005snopt}\cite[\S 2.2]{wachter2006ipopt}.

Such ideas can also be viewed through the lens of optimization on manifolds.
By rewriting constrained problems in Euclidean space as unconstrained problems on manifolds, it is possible to achieve significant empirical performance improvements~\cite{absil2008optimization,boumal2011rtrmc,da2025thorough}.
However, the study of optimization on manifolds with additional non-manifold constraints is still in its infancy~\cite{hauswirth2016riemannianpgd,bergmann2019riemanniankkt,liu2020simple,obara2022riemanniansqp,teng2025riemannian}, so it is not immediately obvious that these benefits will transfer to our problem of interest.

\textbf{Statement of contributions:}
In summary, we present
\begin{enumerate}
    \item a reformulation of a broad class of IK optimization problems, so that the optimizer can take advantage of a given analytic IK solution, and
    \item a detailed comparison between our new formulation and the original formulation across three popular solvers (representing three main paradigms for constrained nonlinear optimization), as well as baseline algorithms, across a variety of experiments. These experiments include simple cases which admit detailed analysis and complex cases that reflect real-world robotics tasks.
\end{enumerate}
Besides these core contributions, we also present
\begin{enumerate}
    \setcounter{enumi}{2}
    \item improvements to analytic IK solutions for the PR2 bimanual mobile manipulator and Hubo humanoid robot (details in \Cref{appx:ik_implementation:hubo,appx:ik_implementation:pr2}),
    \item an integration of Global-IK~\cite{dai2019global} (which we use as a baseline) with modern algorithms for segmenting free space for collision avoidance~\cite{werner2024approximating}, along with an improved objective function (details in \Cref{appx:micp_ik_implementation}), and
    \item an inequality-only formulation of the static stability constraint for a humanoid (or general) robot, yielding a feasible set which is positive-measure in the decision variables (details in \Cref{appx:experiment_implementation:hubo}).
\end{enumerate}
Our results decisively demonstrate that the new formulation achieves high success rates while remaining realtime-capable.
In particular, the success rate for the new formulation is higher than the old for every solver tested and for every experiment, with the largest improvements appearing on challenging collision-free inverse kinematics problems.
These solvers represent three main paradigms for constrained nonlinear optimization (interior point, augmented Lagrangian, and sequential quadratic programming).
\textbf{Altogether, these results strongly suggest that the new formulation makes it fundamentally easier for optimizers to find feasible solutions.}

The remainder of the paper is organized as follows.
\Cref{sec:related_work} contains a detailed overview of existing approaches for solving IK problems.
\Cref{sec:background} covers some theoretical and practical aspects of analytic IK which are relevant to our work, and \Cref{sec:methodology} describes our methodology.
We describe each experimental setup and the corresponding results in \Cref{sec:experiments} and include a brief concluding discussion in \Cref{sec:discussion}.
Discussion of the improvements to existing analytic IK solutions and the global inverse kinematics method are deferred to \Cref{appx:ik_implementation,appx:micp_ik_implementation}, and precise experiment implementation details are deferred to \Cref{appx:experiment_implementation}.

\section{Related Work}
\label{sec:related_work}
Research into analytic IK methods largely began with Pieper's seminal thesis, which showed that the last three joint axes intersecting guarantees an analytic solution for a 6R arm (an arm with six revolute joints)~\cite{pieper1969kinematics}.
The last three joints can be abstracted as a ``spherical'' wrist joint, and then solving the position and orientation of the end-effector can be decoupled.
This solution can also be extended to handle robots with three parallel axes, by interpreting the axes as intersecting at a point at infinity.

Besides such general cases, there are countless papers deriving solutions for specific robot arms or kinematic structures.
In the modern era, when a new robotic arm becomes popular on the market, a specialized analytic IK solution will usually be derived and published soon after.
Popular robots with analytic IK solutions include the KUKA iiwa~\cite{faria2018iiwa}, Franka Emika Panda~\cite{tittel2021frankapanda,he2021frankapanda}, the UR series of robot arms~\cite{hawkins2013universalrobotics}, the Barrett Whole Arm Manipulator~\cite{singh2010barrettwam}, and the arms of a PR2 bimanual mobile manipulator~\cite{ramezani2015pr2}.
Solutions have also been presented for the arms and legs of humanoid robots, such as the Hubo~\cite{ali2010hubo,park2012hubo,oflaherty2013hubo}, NAO~\cite{kofinas2015nao}, and Atlas~\cite{du2014atlas}.

As the process of hand-designing an analytic IK solution is tedious and error-prone, there has been strong interest in automating this process via higher-level algorithms and code generation.
Classical AI approaches including pattern-matching and rule-based methods~\cite{herrera1988symbolic,wenz2007solving}, but these tools have not been widely adopted.
A prominent tool for automatically constructing analytic IK solutions is IKFast~\cite{diankov2010ikfast}, which relies on symbolic manipulation to simplify the equations.
The more recent method IKBT~\cite{zhang2019ikbt} also uses symbolic manipulation, but in the form of a behavior tree, where the system intelligently decides which solutions and simplifications to use.

Another strategy for automatically generating IK solutions is via a decomposition into geometric subproblems.
Building upon many prior works, including the canonical Paden-Kahan subproblems~\cite[\S 3.2]{murray2017mathematical}, IK-Geo~\cite{elias2025ikgeo} showed that all 6R robots can be solved by 6 canonical subproblems, although two of the subproblems do not have closed-form solutions, hence requiring 1D or 2D search.
EAIK~\cite{ostermeier2024eaik} can be used to simplify robot descriptions so that the closed-form subproblems of IK-Geo can be applied.

Advances in the field of numerical algebraic geometry have also been applied towards solving IK problems \cite{wampler2011numerical}.
Such techniques are readily enabled by converting the trigonometry in the IK map into a system of polynomial equations through a variety of substitutions including the tangent half-angle substitution~\cite{raghavan1990kinematic} (also known as the stereographic projection~\cite[\S 6.4.3]{russtedrake2024manipulation}), conversion to a Chebyshev polynomial~\cite[\S 2.8]{dumitrescu2007positive}, or the algebraic embedding $\sin(x) \mapsto s$, $\cos(x)\mapsto c$, $s^{2}+c^{2}=1$ \cite[\S 3.1.1]{wampler2011numerical}.
The resulting system of polynomial equations can be solved with polynomial elimination, homotopy continuation, and Gr\"obner basis methods~\cite{raghavan1995solving}.

Polynomial elimination has been particularly useful for IK, since Raghavan and Roth~\cite{raghavan1990kinematic,raghavan1993inverse} designed a polynomial elimination strategy for the generic 6R IK problem.
This procedure yields a univariate polynomial of degree 16, for which each root corresponds to an IK solution.
Followup work by Lee et al.~\cite{lee1991complete}, Kohli and Osvatic~\cite{kohli1993inverse}, Manocha and Canny~\cite{manocha2002efficient}, and Husty et al.~\cite{husty2007new} has resulted in simpler, more efficient, and more numerically stable algorithms for the general 6R problem.
Furthermore, such approaches elegantly handle the question of reachability, as nonreachable end-effector poses yield complex solutions to the polynomial system.

In contrast to the large number of specialized solutions in the analytic IK literature, the optimization-based IK literature transcribes the IK problem as a generic nonlinear optimization problem, and then applies powerful off-the-shelf solvers such as IPOPT~\cite{wachter2006ipopt} or SNOPT~\cite{gill2005snopt}.
For example, one can use sequential quadratic programming to solve complex, whole-body kinematics problems~\cite{dai2014whole,fallon2015architecture,beeson2015trac}.
Since these optimization formulations are almost always nonconvex, there has been significant research on clever initialization strategies to aid the solver~\cite{vahrenkamp2015ikmap,yuan2025iksel}.
Massive GPU parallelism can also be brought to bear, as in the cuRobo motion generation pipeline~\cite{sundaralingam2023curobo} and HJCD-IK~\cite{yasutake2025hjcd}.
Alternatively, there are various constraint relaxation strategies which can produce global solutions.
Porta et al. used a branch-and-prune global optimization strategy~\cite{porta2005branch}, together with linear constraint relaxations~\cite{porta2009linear}.
Dai et al.~\cite{dai2019global} reduced all nonlinearities into bilinearities, and used McCormick envelopes~\cite{mccormick1976envelopes} to describe a mixed-integer convex relaxation.
More recent optimization formulations have used semidefinite relaxations~\cite{giamou2022sdp,wu2023sdp}, polynomial optimization~\cite{maric2020sos,trutman2022globally}, and Riemannian optimization~\cite{maric2021riemannian}.

Although we do not directly compare against them in this paper, it is important to mention the rich variety of specialized numerical IK algorithms.
Canonical numerical approaches include Jacobian psuedoinverse, Jacobian transpose, and damped least squares~\cite{buss2004introduction}.
Forward and Backward Reaching Inverse Kinematics (FABRIK) has been especially popular in the computer graphics community~\cite{aristidou2011fabrik,aristidou2016fabrik}, thanks to its fast convergence and visually realistic results.
And finally, machine learning has also served as a powerful tool for inverse kinematics.
Given solutions to the IK problem for specific configurations of a robot (generally computed by simply solving forward kinematics, and pairing the joint angles with the resulting end-effector pose), learning a general IK functions can be cast as a regression problem and solved via neural networks~\cite{oyama2001neuralnetwork,ardizzone2019neuralnetwork,kim2021neuralnetwork}, or as a generative modeling problem~\cite{ames2022neuralnetwork,ho2023vae,zhang2025diffusion}.
However, these methods will generally experience poorer accuracy than other approaches.

Although analytic and numeric IK algorithms each have been studied in great detail, there have been relatively few works on hybrid approaches that combine the two.
These approaches have generally been specialized to specific manipulators and/or numerical algorithms, whereas our approach is designed to be used with any analytic IK solution and any optimization algorithm.
Tolani et al.~\cite{tolani2000real} describe how a broader class of end-effector constraints (e.g. position and partial orientation, or aiming only) can still be handled using analytic IK for a 7DoF arm with spherical shoulder and wrist, falling back to a generic optimization approach when a solution cannot be found.
Ananthanarayanan and Ord\'o\~nez~\cite{ananthanarayanan2015real} use FABRIK and a custom optimization-based redundancy resolution strategy to solve IK problems for robot arms with an odd number of joints and a spherical wrist.
Jin et al.~\cite{jin2020efficient} use derivative-free optimization together with analytic IK for a space station remote manipulator system~\cite{laryssa2002spacemanipulator}.

Finally, analytic and learning approaches have also been combined to achieve favorable results.
Nguyen and Marvel~\cite{nguyen2022modeling} treat an analytic IK solution as the prior mean distribution for a Gaussian process.
Li et al.~\cite{li2021hybrik} use IK as part of a human pose and shape esimation pipeline, where the state of each joint is estimated using an analytic solution and a neural network.

\section{Background}
\label{sec:background}
Independent of solving IK, researchers have studied the underlying topology of robot kinematics in great detail.
We focus on the 3D case (6DoF goal), as the 2D case is much simpler.
We briefly discuss some key results that are relevant to our work, before presenting a more practical discussion about the internal structure of an analytic IK function.
But first, we introduce some basic notation.

We write $[I]=\set{1,\ldots,I}$, and use $\mb T^d$ to denote the $d$-torus.
To represent changes of coordinates in Euclidean space, we use monogram notation~\cite[\S 3.1]{russtedrake2024manipulation}.
$\monogram{X}{A}{B}\in\SE(3)$ is the pose of frame $B$ relative to (and measured in) frame $A$, expressed as a $4\times 4$ homogeneous transformation matrix.
When this pose is a function of variables $q$, we write $\monogram{X}{A}{B}(q)$.
For example, the pose of the gripper frame $G$ on the end of a robot arm in the world coordinate frame $W$, when the robot's joint angles are $q$, is written $\monogram{X}{W}{G}(q)$.

The position of frame $B$ relative to (and expressed in) frame $A$ is denoted $\monogram{p}{A}{B}$.
The orientation of frame $B$ relative to frame $A$ (represented with Euler angles) is denoted $\monogram{o}{A}{B}$.
We can write the homogeneous transformation matrix described by such a position and orientation as $X(\monogram{p}{A}{B},\monogram{o}{A}{B})$.
Similarly, given $\monogram{X}{A}{B}$, we can obtain its position with $p(\monogram{X}{A}{B})$, and the set of Euler angles representing its orientation with $o(\monogram{X}{A}{B})$.
Since the Euler angles surject onto $\SO(3)$, we can always select a member from $o(\monogram{X}{A}{B})$.

\subsection{The Topology of Robot Kinematics}
\label{sec:background:kinematic_topology}

Suppose we have a robot arm with $d$ revolute joints.
Then its configuration space is $\mb T^d$, or some subset if there are joint limits.
As a function of the configuration $q\in\mb T^d$, the end-effector pose (frame $G$) relative to the base of the arm (frame $B$) is $\monogram{X}{B}{G}(q)\in\SE(3)$ (or $\SE(2)$ for a 2D robot).
This function is known as \emph{forward kinematics}, and its image is called the \emph{reachable set}.

The \emph{kinematic Jacobian} $D\monogram{X}{B}{G}(q)\in\R^{6\times d}$ (or $\R^{3\times d}$ for a 2D robot) maps a velocity in joint space to the corresponding spatial velocity of the end-effector.
A \emph{kinematic singularity} is a point $q$ for which $D\monogram{X}{B}{G}(q)$ does not have full row-rank; for a robot arm, this indicates there is some direction (in position and/or orientation) in which the end-effector cannot instantaneously be moved.
Although almost all robot arms have singularities somewhere, they can generally be avoided for robots with \emph{kinematic redundancy}: more than 6 DoF (or more than 3 DoF for a 2D workspace)~\cite{hollerbach1985optimum,gottlieb1988topology}.
However, the set of end-effector poses only achievable by singular configurations is measure-zero within the reachable workspace~\cite{sard1942measure}, and our approach does not encounter them in practice.

A kinematically-redundant robot arm can move continuously while keeping its end-effector fixed; this is called \emph{self-motion}, and has been extensively studied.
Away from singularities, the self-motion manifolds are themselves tori, the number of self-motion manifolds is never more than $16$, and the number of disjoint self-motion manifolds generally decreases as the degree of kinematic redundancy is increased~\cite{burdick1989inverse,luck1993self}.

\subsection{A Look Inside Analytic IK Functions}
\label{sec:background:inside_analytic_ik}

In this subsection, we take a look at the internal structure of analytic IK mappings.
An analytic IK function takes in three arguments:
\begin{enumerate}
    \item the target end-effector pose in $\SE(3)$ (or $\SE(2)$ for a 2D workspace),
    \item continuous self-motion parameters $\psi\in\Psi$, describing the point within a self-motion manifold, and
    \item discrete self-motion parameters $\kappa\in\ms K$, describing which self-motion manifold is being considered.
\end{enumerate}
The output of the function is a set of joint angles for the arm.
So explicitly, an analytic IK function is denoted by
\begin{equation}
    \label{eq:ik_mapping}
    \IK:\SE(3)\times\Psi\times\ms{K}\to\R^d,
\end{equation}
where we have used $\ms K$ to denote the set of self-motion manifolds.

The range of an analytic IK function typically does not match the true set of reachable configurations.
Analytic IK functions generally do not consider joint limits, so the output of the function must be checked (although one can sometimes describe subsets of the self-motion manifold for which the joint limits are respected~\cite{faria2018iiwa}).
A bigger problem is due to the limited domain, since the set of reachable configurations of a robot arm is generally not all of $\SE(3)$.
This reachability issue appears within the mapping as domain-restricted functions like $\arccos$, e.g.,~\cite[Eq. (18)]{faria2018iiwa} -- nonreachable end-effector poses will cause domain violations of these functions.

Besides issues of joint limits and reachability, singularities are also a major concern.
Typically, an IK mapping will be numerically unstable near kinematic singularities, as even a slight perturbation would be nonreachable (or require a significant reconfiguration to achieve the resulting motion).
Additionally, there is another class of singularities, called representational singularities.
They appear for the same reason that kinematic singularities appear: the domain of the forward kinematics mapping is toroidal, but the target space is $\SE(3)$, which is topologically distinct.
Refer to Elias and Wen~\cite{elias2025ikgeo} for further discussion.
Sometimes, some of these singularities can be avoided by defining the mappings such that they occur outside of the joint limits~\cite{elias2024stereographicsew}.

\section{Methodology}
\label{sec:methodology}
In this section, we will describe the general form of the optimization problems of interest and how they can be reformulated using analytic IK.
As in \Cref{sec:background}, we focus on the case of a 6DoF goal in a 3D workspace, although a 2D workspace can be handled as well by using $\SE(2)$ instead.
For each experiment, we are given a robotic system with $d$ joints, along with an analytic inverse kinematic mapping \eqref{eq:ik_mapping}.
For simplicity, we state our problem formulation in terms of a single kinematic chain, although there is a natural generalization to general kinematic trees that we will discuss later.
We represent elements of $\SE(3)$ as $4\times 4$ homogeneous transformation matrices and elements of $\ms K$ as a tuple of binary variables.

To handle the domain restrictions, we require that $\IK$ is constructed to clip its inputs for domain-limited functions like $\arccos$.
This means that for a nonreachable configuration, the joint angles output by $\IK$ will not actually achieve the specified end-effector pose.
Thus, we construct \emph{probing functions}
\begin{equation}
    \label{eq:ik_probe}
    \mc D_k:\SE(3)\times\Psi\times\ms{K}\to\R
\end{equation}
that return intermediate values from within the computations of $\IK$ before a clipping operation, such that if $\mc D_k(\monogram{X}{W}{G},\psi,\kappa)\ge 0$ for all $k\in[K]$, then the clipping within $\IK$ has no effect.
For example, if $\IK$ requires computing $\arccos(t)$ for some intermediate quantity $t=f(\monogram{X}{W}{G},\psi,\kappa)$, we would introduce the probing functions
\begin{subequations}
\label{eq:probing_functions_example}
\begin{align}
    & \mc D_1:(\monogram{X}{W}{G},\psi,\kappa)\mapsto 1-f(\monogram{X}{W}{G},\psi,\kappa),\\
    & \mc D_{2}:(\monogram{X}{W}{G},\psi,\kappa)\mapsto 1+f(\monogram{X}{W}{G},\psi,\kappa).
\end{align}
\end{subequations}
If $\mc D_k(\monogram{X}{W}{G},\psi,\kappa)\ge 0$ for each $k\in[K]$, then $\monogram{X}{W}{G}$ is a reachable configuration.

Imposing this \emph{reachability constraint} in our optimization problems is essential for convergence.
Without it, if the optimizer loses feasibility, then the analytic IK function is undefined, and no gradient information is available to recover.

Let $q_{\lb},q_{\ub}\in(\R\cup\set{\pm\infty})^d$ be the lower and upper joint limits, and $\monogram{X}{W}{G}_{\des}$ the desired configuration of the end-effector.
Finally, we have generic functions $f,g_i,h_j:\R^d\to\R$ for $i\in[I]$ and $j\in[J]$ to represent the cost, inequality constraints, and equality constraints (respectively).

\subsection{Reformulation of Optimization IK with Analytic IK}
\label{sec:methodology:formulation}

We begin with the standard formulation for an IK optimization problem, which can already be solved using generic nonlinear optimization methods.
We then present a reformulation of the problem, where we use analytic IK as a change of coordinates.

\subsubsection{An Existing Formulation}
\label{sec:methodology:formulation:old}

The standard formulation for an optimization inverse kinematics problem is
\begin{subequations}
\label{eq:old_formulation}
\begin{align}
        \min_q \colsep & f(q)
            \label{eq:old_formulation:cost}\\
        \ST \colsep & q\in\R^d
            \label{eq:old_formulation:variables}\\
        \colsep & \monogram{X}{W}{G}(q)=\monogram{X}{W}{G}_{\des}
            \label{eq:old_formulation:ik_constraint}\\
        \colsep & q_{\lb}\le q\le q_{\ub},
            \label{eq:old_formulation:joint_limits}\\
        \colsep & g_i(q)\le 0, & \forall i\in [I],
            \label{eq:old_formulation:generic_inequality}\\
        \colsep & h_j(q)=0, & \forall j\in [J].
            \label{eq:old_formulation:generic_equality}
\end{align}
\end{subequations}
Here, the joint angles are the decision variables, and the underlying ``inverse kinematics'' constraint is encoded in \eqref{eq:old_formulation:ik_constraint}.
$f$ is a user-defined cost function -- joint-centering is a common choice.
$g_i$ and $h_j$ are generic additional constraints, such as collision-avoidance or humanoid stability.

\subsubsection{Reformulation}
\label{sec:methodology:formulation:new}

The variable $q$ in \eqref{eq:old_formulation} and constraint \eqref{eq:old_formulation:ik_constraint} are readily eliminated by the analytic IK function \eqref{eq:ik_mapping}.
This results in an optimization over the end-effector position $p$ and orientation $o$, the continuous self-motion parameters $\psi \in \Psi$, and the discrete self-motion parameters $\kappa \in \ms{K}$,
\begin{subequations}
\label{eq:new_formulation}
\begin{align}
        \min_{p,o,\psi,\kappa} \colsep & f\left(\IK(X(p,o),\psi,\kappa)\right)
            \label{eq:new_formulation:cost}\\
        \ST \colsep & p\in\R^3,o\in\mb T^3,\psi\in\Psi,\kappa\in\ms K,
            \label{eq:new_formulation:variables}\\
        \colsep & p=\monogram{p}{W}{G}_{\des}, \; o=\monogram{o}{W}{G}_{\des},
            \label{eq:new_formulation:ik_constraint}\\
        \colsep & q_{\lb}\le \IK(X(p,o),\psi,\kappa)\le q_{\ub},
            \label{eq:new_formulation:joint_limits}\\
        \colsep & g_i\left(\IK(X(p,o),\psi,\kappa)\right)\le 0 & \forall i\in [I],
            \label{eq:new_formulation:generic_inequality}\\
        \colsep & h_j\left(\IK(X(p,o),\psi,\kappa)\right)=0 & \forall j\in [J],
            \label{eq:new_formulation:generic_equality}\\
        \colsep & D_k(p,o,\psi)\ge 0 & \forall k\in [K],
            \label{eq:new_formulation:reachable}
\end{align}
\end{subequations}
To solve this mixed-integer nonconvex optimization problem, we fix a choice of $\kappa_0\in\ms K$.
(Since there are few choices for $\kappa_0$, one could easily solve the problem for each possible choice.)
The reachability constraint \eqref{eq:new_formulation:reachable} requires that the joint angles output from the IK function actually achieve the expected gripper pose; see \Cref{sec:background:inside_analytic_ik} for further details.
As the intent of our approach is to remove nonlinear equality constraints, we represent orientation with Euler angles, although one could also use exponential coordinates.
Unit quaternions or full rotation matrices would be unsuitable, as they are described by nonlinear equality constraints.

We draw the reader's attention to several key aspects of this optimization problem.
The most important part is the elimination of the nonlinear equality constraint \eqref{eq:old_formulation:ik_constraint} -- the only remaining nonlinear equality constraints are the additional user-specified ones.
Most solvers will be able to satisfy the linear constraints at every step of the algorithm, or they are guaranteed to satisfy the linear constraints at final convergence up to the accuracy of a linear system solve.
This ensures feasibility of the IK manifold constraint to a very precise tolerance.

When there are additional nonlinear equality constraints, the performance difference between the two formulations is less clear, and we cannot guarantee the same tight tolerances.
In this setting, later empirical evidence will suggest we still match the performance of the old formulation in runtime and success rate (see \Cref{sec:experiments:hubo}).

\subsubsection{Some Immediate Generalizations}
\label{sec:methodology:formulation:generalizations}

Sometimes, we may not want to entirely constrain the pose of the end-effector.
For example, we may fix the position but allow the orientation to vary, or we could allow some bounded amount of error in the end-effector pose.
We might also want to remove the constraint outright, and instead just impose a cost on the disparity between the achieved and desired end-effector pose.
All of these possibilities can be easily handled in the old formulation, by appropriately modifying the constraint \eqref{eq:old_formulation:ik_constraint}, and in the new formulation, by treating $p$ and $o$ from \eqref{eq:new_formulation:ik_constraint} as decision variables.
Experiments involving such broader constraints are presented in \Cref{sec:experiments:grasp_selection,sec:experiments:hubo}.

Another common way to generalize the IK problem is to simultaneously consider multiple kinematic chains.
In this case, we may be given separate analytic IK functions for each kinematic chain, but the two chains will be coupled in some way.
For example, we may wish for a bimanual robot to simultaneously achieve a desired end-effector pose for both of its arms, as shown in \Cref{sec:experiments:pr2}.
In this case, we leverage a separate analytic IK solution for each arm, which takes in the poses of the end-effector target and the base of the arm.
The pose of the base of the arm, in turn, can be written as a function of the pose of the robot base, and any intermediate joints (the torso lift joint, in the case of the PR2).
Another example is grasping an object with a humanoid robot while fixing its feet to be flat on the ground (shown in \Cref{sec:experiments:hubo}).
In this case, we have analytic IK solutions for both arms and both legs, and the base of each arm and leg is determined as a function of the pose of the torso and the hip rotation angle.

\subsubsection{Handling Domain Limitations of the IK Function}
\label{sec:methodology:formulation:ik_domain_limitations}

As discussed in \Cref{sec:background:inside_analytic_ik}, the analytic IK mappings are undefined for nonreachable configurations.
Most optimizers do not guarantee that their intermediate iterates will be feasible, even when the initial guess is feasible.
Therefore, special care must be taken to ensure the solver does not have errors (e.g., due to taking the $\arccos$ of a number larger than one).

For most of our experiments, we take a combined strategy of restricting the domain of the inputs with the probing functions, and then projecting the intermediate values onto the appropriate domains before applying operations like $\arccos$.
This can be interpreted as extending the domain of the analytic IK function to return the nearest valid solution.
We use nearest in an imprecise sense, as opposed to, for example, the least squares solutions produced by IK-Geo~\cite{elias2025ikgeo}.
Although the clipping operation loses gradient information, the reachability constraints provide gradient information from before this step, guiding the optimizer to reachable configurations.
In practice, we clip to slightly inside the domain limits, ensuring finite gradients.
This strategy was effective for all of our experiments, except the humanoid stability experiments described in \Cref{sec:experiments:hubo}.
Refer to \Cref{appx:experiment_implementation:hubo} for details.

\section{Experiments}
\label{sec:experiments}

\begin{figure*}
    \centering

    \setlength\tabcolsep{0.1em}

    \begin{tabular}{cccc}
        \begin{subfigure}[b]{0.28\linewidth}
            \centering
            \includegraphics[height=3cm]{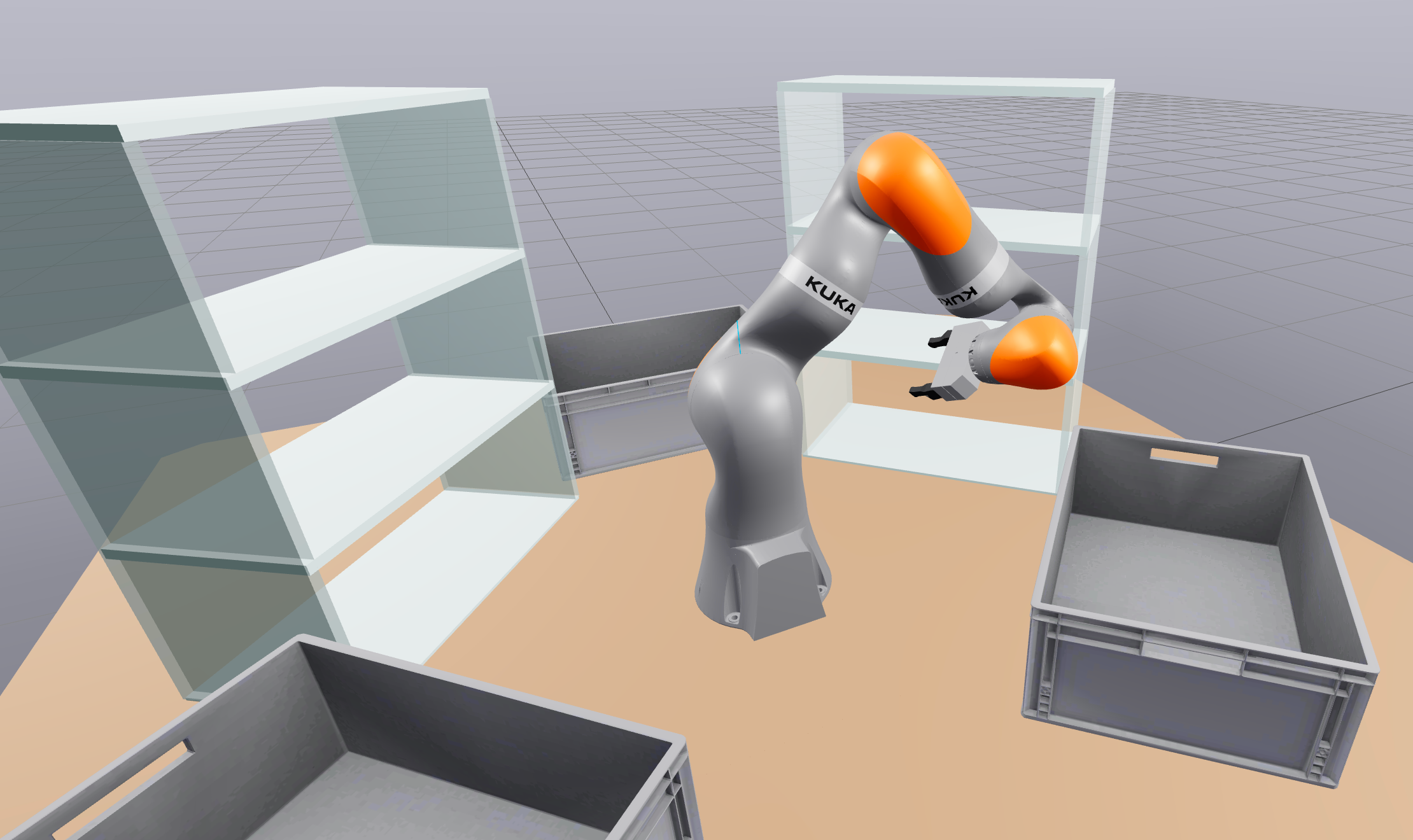}
            \caption{Arm on a Table}
            \label{fig:experimental_setup:arm_on_a_table}
        \end{subfigure}
        &
        \begin{subfigure}[b]{0.28\linewidth}
            \centering
            \includegraphics[height=3cm]{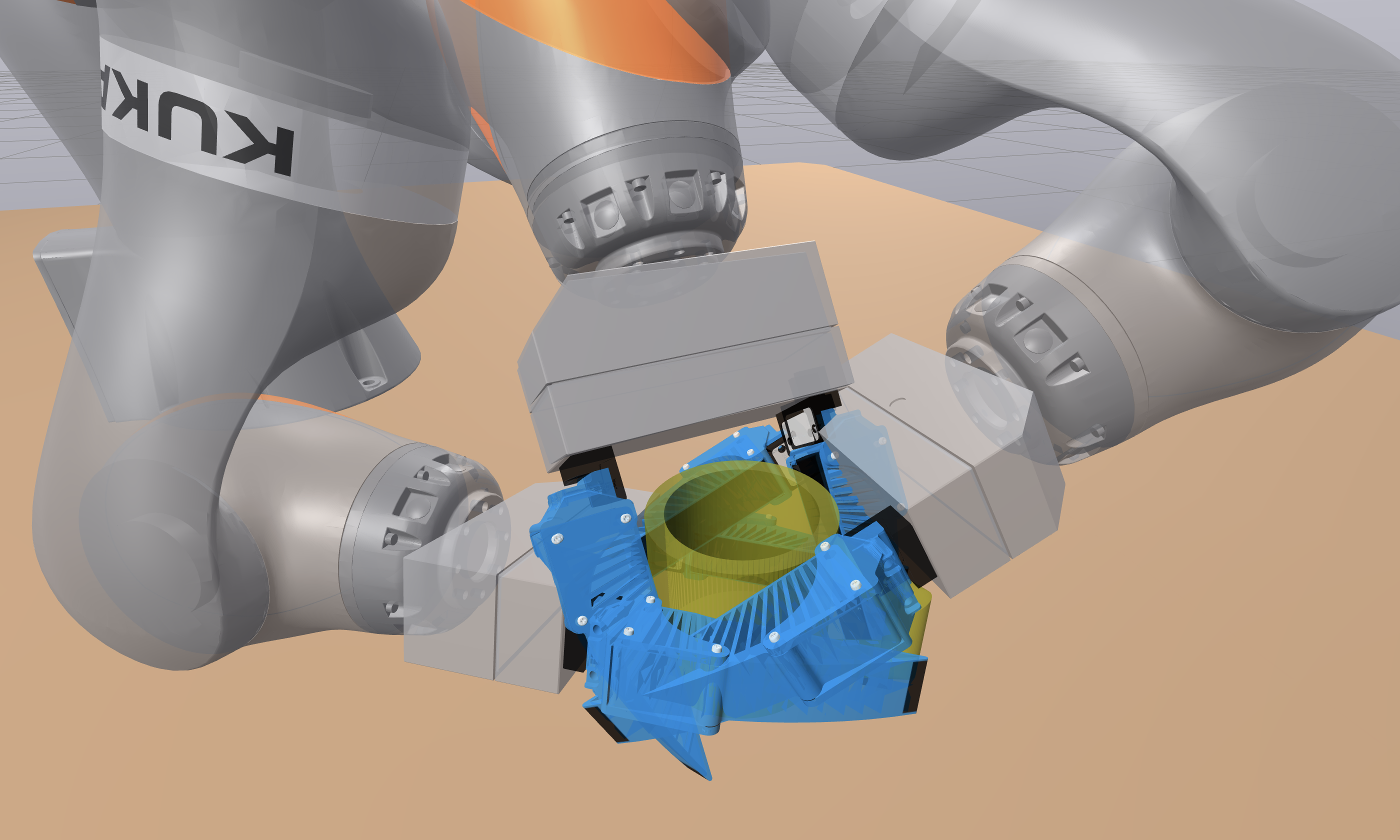}
            \caption{Grasp Selection (Shelves Hidden)}
            \label{fig:experimental_setup:grasp_selection}
        \end{subfigure}
        &
        {} \\  %
        &
        {} \\  %

        \begin{subfigure}[b]{0.28\linewidth}
            \centering
            \includegraphics[height=3cm]{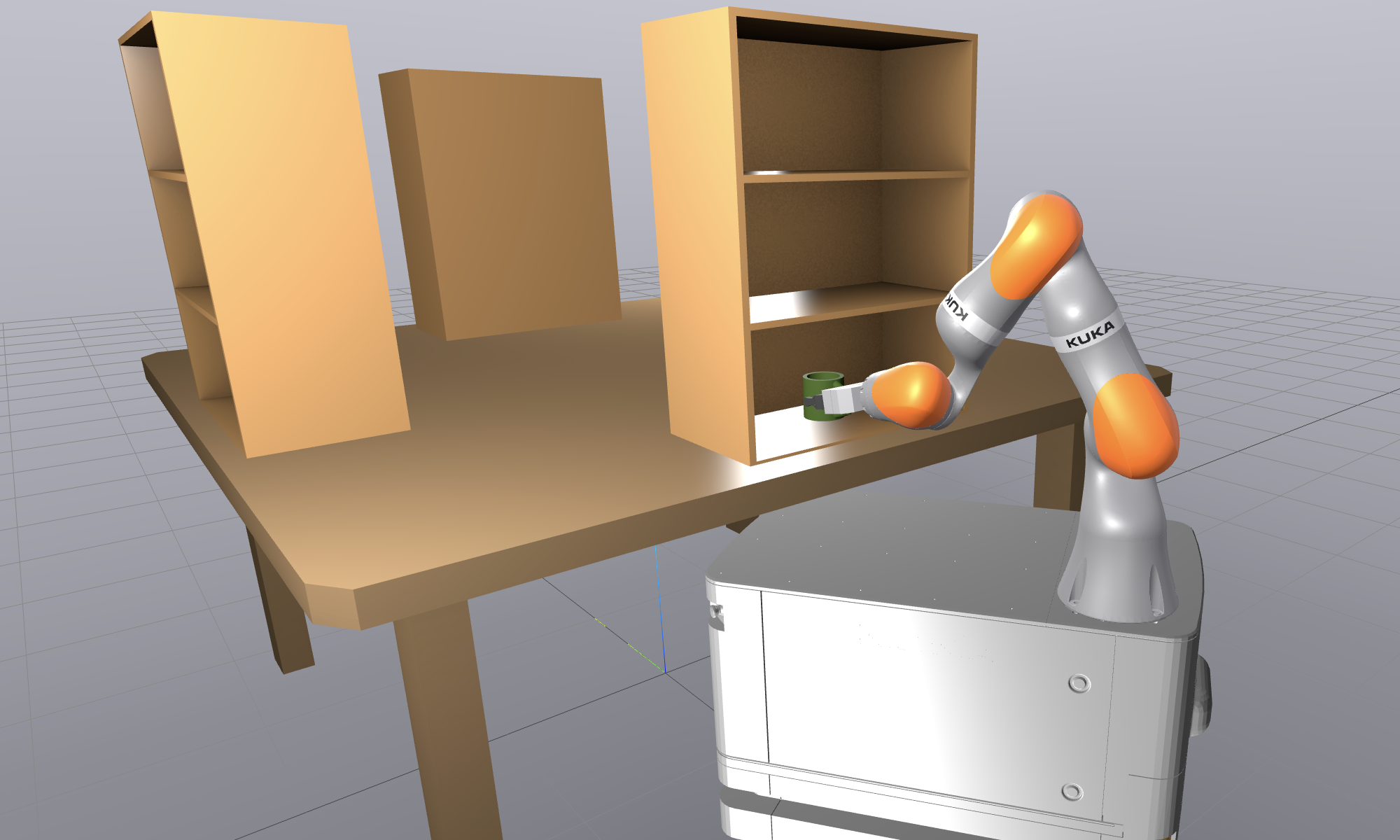}
            \caption{Mobile Manipulator}
            \label{fig:experimental_setup:kmr_iiwa}
        \end{subfigure}
        &
        \begin{subfigure}[b]{0.28\linewidth}
            \centering
            \includegraphics[height=3cm]{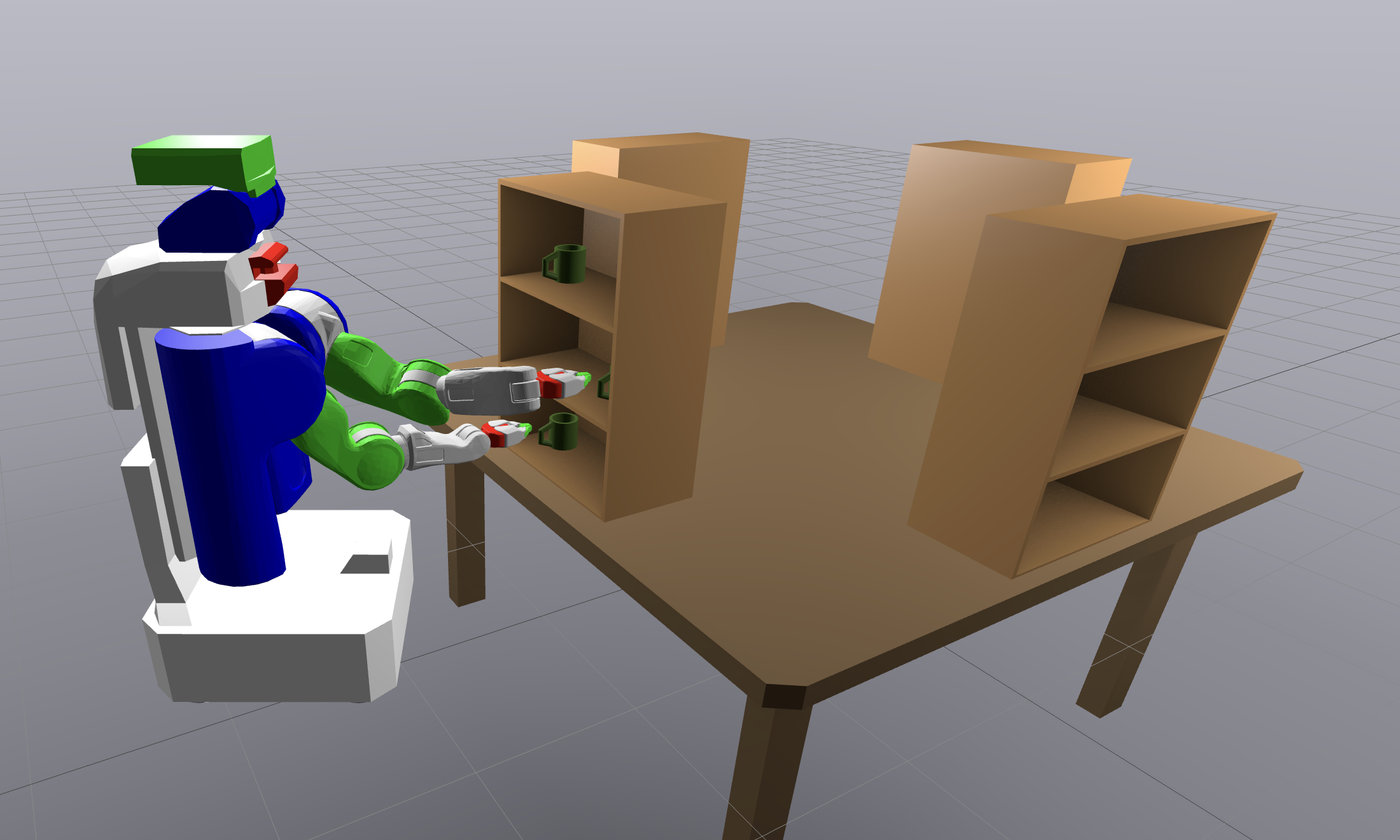}
            \caption{Bimanual Mobile Manipulator}
            \label{fig:experimental_setup:pr2}
        \end{subfigure}
        &
        \begin{subfigure}[b]{0.2\linewidth}  %
            \centering
            \smash[t]{\includegraphics[height=7.1cm]{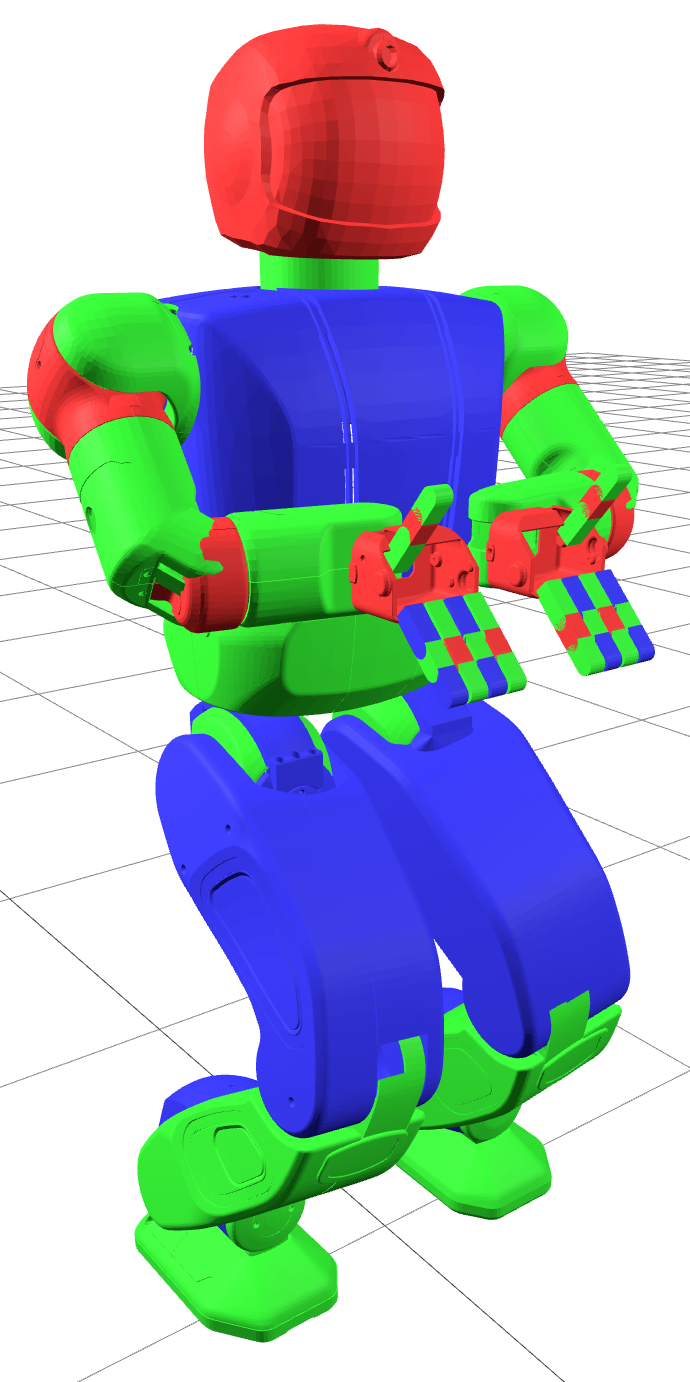}}
            \caption{Humanoid Stability}
            \label{fig:experimental_setup:hubo}
        \end{subfigure}
        &
        \begin{subfigure}[b]{0.2\linewidth}  %
            \centering
            \smash[t]{\includegraphics[height=7.1cm]{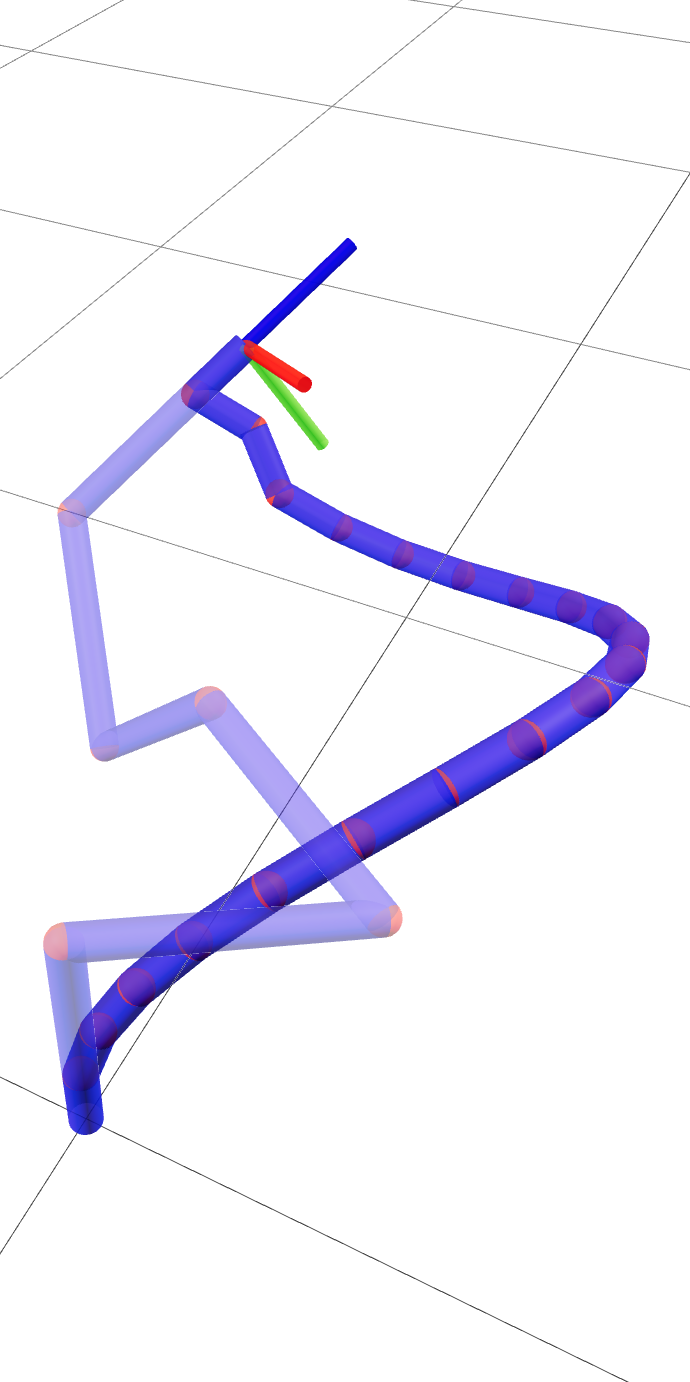}}
            \caption{Scaling}
            \label{fig:experimental_setup:scaling}
        \end{subfigure}
    \end{tabular}
    \caption{Setup for our robot simulation experiments. (b) shows three poses which are feasible for one instance of the grasp selection experiment, with the shelves hidden for better visibility. (e) shows the nominal stable configuration used in the Hubo experiments. (f) shows the 3d scaling setup, where arms with different numbers of joints are used for the same IK problem.}
    \label{fig:experimental_setup}
\end{figure*}

\begin{table*}
    \renewcommand{\arraystretch}{1.25}
    \setlength{\tabcolsep}{3pt}
    \centering
    \begin{tabularx}{\textwidth}{|c|c|c|VC|C|VC|C|VC|C|V>{\centering\arraybackslash}p{0.8cm}|>{\centering\arraybackslash}p{0.8cm}|Vc|} \hline
        \multirow{3}{*}{Experiment} &
        \multirow{3}{*}{$d$} &
        \multirow{3}{*}{$d'$} &
        \multicolumn{2}{c|V}{\multirow{2}{*}{\parbox{2.8cm}{\centering Interior Point\\(IPOPT)}}} &
        \multicolumn{2}{c|V}{\multirow{2}{*}{\parbox{2.8cm}{\centering Augmented Lagrangian\\(NLOPT)}}} &
        \multicolumn{2}{c|V}{\multirow{2}{*}{\parbox{2.8cm}{\centering Sequential Quadratic\\Programming (SNOPT)}}} &
        \multicolumn{2}{c|V}{\multirow{2}{*}{\parbox{1.6cm}{\centering Global-IK}}} &
        \multirow{3}{*}{Sampling}
        \\

        & & & \multicolumn{2}{c|V}{} & \multicolumn{2}{c|V}{} & \multicolumn{2}{c|V}{} & \multicolumn{2}{c|V}{} &
        \\ \cline{4-11}
        & & & New & Old & New & Old & New & Old & Fast & Precise &
            \\ \hline
        Arm on a Table & 7 & 1 & \bf 0.8951 & 0.5392 & \bf 0.8733 & 0.049 & \bf 0.9144 & 0.463 & 0.88 & 0.97 & 1.00 \goldstar
            \\ \hline
        Grasp Selection & 7 & 5 & \bf 0.9893 \goldstar & 0.9875 & \bf 0.47 & \bf 0.47 & \bf 0.7344 & 0.6625 & \tableNA & \tableNA & 0.525
            \\ \hline
        Mobile Manipulator & 10 & 4 & \bf 0.7182 & 0.5946 & \bf 0.405 & 0.0074 & \bf 0.368 & 0.2884 & \tableNA & \tableNA & 0.95 \goldstar
            \\ \hline
        Bimanual Mobile Manipulator & 18 & 6 & \bf 0.7819 \goldstar & 0.5243 & \bf 0.68 & \tableNA & \bf 0.3289 & 0.2466 & \tableNA & \tableNA & 0.32
            \\ \hline
        Humanoid Stability (Inequality) & 31 & 19 & \bf 0.97 \goldstar & 0.70 & \bf 0.67 & \tableNA & \tableNA & \tableNA & \tableNA & \tableNA & \tableNA
            \\ \hline
        Humanoid Stability (Equality) & 31 & 19 & \bf 0.96 \goldstar & \bf 0.96 \goldstar & \bf 0.46 & \tableNA & \tableNA & \tableNA & \tableNA & \tableNA & \tableNA
            \\ \hline
    \end{tabularx}
    \caption{
        Success rates for each experiment, where any feasible solution is treated as a success.
        For each solver, the formulation with the higher success rate is bolded, and the overall highest success rate has a gold star.
        We also report the dimension of the configuration space (abbreviated $d$) and the dimension of the constraint manifold (abbreviated $d'$).
        Each general nonlinear solver achieves a equal or better success rate with the new formulation than the old in all but one case.
        The Global-IK baseline was only tractable for the first experiment due to runtime issues.
        The sampling baseline is only effective when $d'$ is low and requires careful tuning on a per-experiment basis.
        Details on the exact sampling strategies for each experiment are given in \Cref{appx:experiment_implementation}.
    }
    \label{tab:success_rates}
\end{table*}

When comparing two different nonlinear optimization formulations, comprehensive experiments are needed to clearly show the difference in performance.
Throughout our experiments, we compare our new formulation to the old formulation with three solvers: the interior point solver IPOPT~\cite{wachter2006ipopt}, the sequential quadratic programming solver SNOPT~\cite{gill2005snopt}, and an augmented Lagrangian solver~\cite{conn1991augmentedlagrangian,birgin1991augmentedlagrangian} made available via the NLOPT interface~\cite{johnson2007nlopt}.
As a baseline, we compare with simply drawing random samples from the self-motion manifold and picking the lowest-cost feasible one.

Finally, we compare to Global-IK, the mixed-integer optimization approach of Dai et al.~\cite{dai2019global} for our first experiment.
To handle collision avoidance, this baseline requires an inner approximation of the free space as the union of convex polyhedra.
In practice, Global-IK may fail to find a feasible solution when one exists, due to the gap between this inner approximation and the true free space.
Also, due to the relaxation of $\SO(3)$, it may return a configuration that is in-collision or does not exactly achieve the desired end-effector pose; we only treat collisions as failures when measuring success rate.
Global-IK could not be used for the remaining experiments due to scaling issues or the limited set of constraints it can handle.
To strengthen it as a baseline, we leverage modern approaches for approximating the free space automatically~\cite{werner2024approximating,petersen2023growing,werner2024faster}, as opposed to the axis-aligned bounding boxes used in the original work.
For a fair comparison, we also designed new cost functions, which more accurately correspond to the quadratic joint centering cost used by our nonlinear optimization formulations.
See \Cref{appx:micp_ik_implementation} for more details.

The experimental setups are visualized in \Cref{fig:experimental_setup}, and success rates, optimal costs, and runtimes for these experiments are included in Tables \ref{tab:success_rates}, \ref{tab:optimal_costs}, and \ref{tab:runtimes} (respectively).
We also have experiments comparing the approaches as the degree of self-motion grows; these results are discussed in \Cref{sec:experiments:scaling}.

Some implementation details are deferred to \Cref{appx:experiment_implementation}, including
\begin{itemize}
    \item common costs and constraints not inherently connected to our formulation that appear frequently throughout our experiments,
    \item precise formulations of the optimization problems solved in each experiment,
    \item the exact procedures used for the sampling baseline, and
    \item how the random end-effector target configurations are selected.
\end{itemize}
The implementation of the Global-IK baseline is discussed in \Cref{appx:micp_ik_implementation}.

\subsection{Arm on a Table}
\label{sec:experiments:arm_on_a_table}

Our first experiment is a classic collision-free IK problem: a 7DoF KUKA LBR iiwa 14 arm, mounted in the center of the table, surrounded by several shelves and bins, as shown in \Cref{fig:experimental_setup:arm_on_a_table}.
We constrain the gripper to a specific target pose, demanding $\monogram{X}{W}{G}(q)=\monogram{X}{W}{G}_{\des}$, as well as imposing joint limits, collision-avoidance, and a quadratic joint centering cost.
We selected 100 random targets, and ran the nonlinear solvers on 100 random initial guesses per target.
See \Cref{appx:experiment_implementation:arm_on_a_table} for details.

As expected, the new formulation significantly outperforms the old formulation in success rate.
Since there is only one degree of redundancy given the end-effector pose, the sampling baseline can densely sample the search-space for the redundancy parameter.
The self-motion parameterization is not isometric, so the sampling distribution is uneven in configuration space.
In practice, we found 500 samples per target pose is sufficient to achieve close to the global optimum.

The Global-IK baseline also achieves a very high success rate on both settings, although it requires significant runtime.
In both settings, the Global-IK finds a lower-cost solution than sampling, due to the relaxation of the kinematic constraints~\cite[\S 3.3]{dai2019global}.
In the case of the fast settings, this results in an average of about eight centimeters of error in the end-effector position and about one centimeter for the precise settings.

\begin{figure*}
	\centering
	\includegraphics[width=\textwidth]{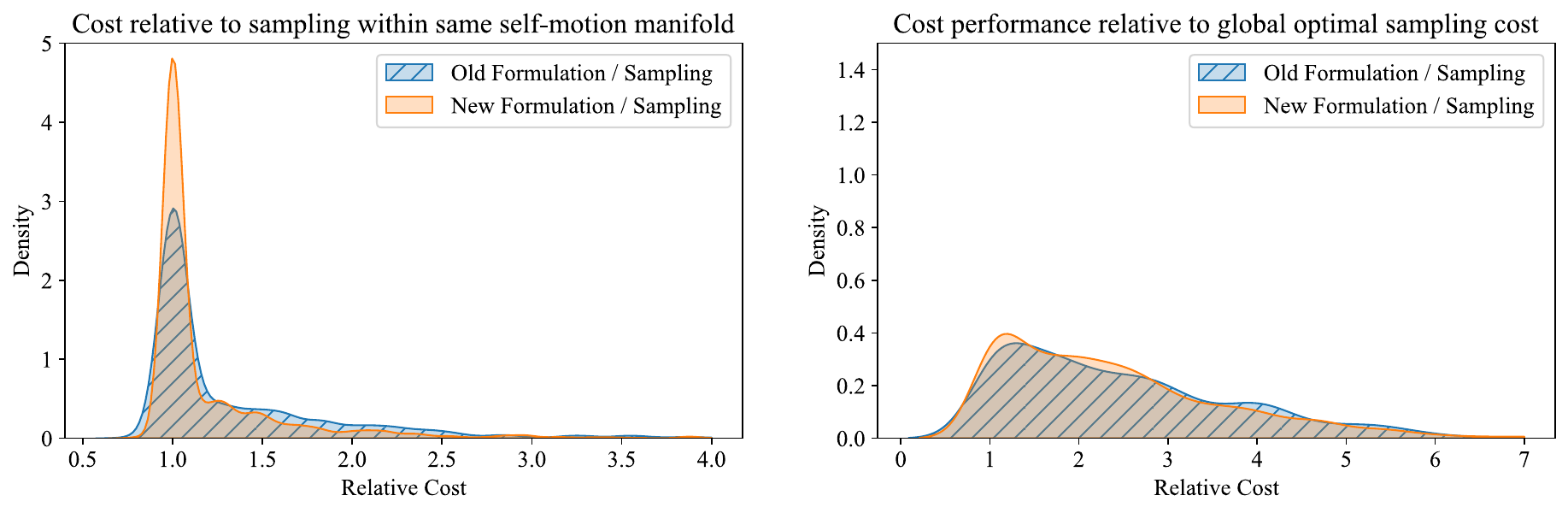}
	\caption{
		Optimal joint centering cost found by new and old formulation divided by the optimal costs found through sampling.
        When a solution is found (with either formulation) in the same self-motion manifold as the global optimum, it is generally that global optimum, with the new formulation obtaining costs close to global optimum more often.
        When the solution is in a different self-motion manifold, both formulations have similar performance.
        The cost difference seen in \Cref{tab:optimal_costs} is due to the new formulation's restriction to producing solutions from a prespecified self-motion manifold (unlike the sampling approach). This difference disappears if the sampling approach is restricted to the same self-motion manifold as the new formulation.
	}
	\label{fig:iiwa_cost_cbundles}
\end{figure*}

We also study how the choice of IK branch affects the solution cost.
We increase the sampling resolution ten-fold to establish global optima as a baseline for comparison in \Cref{fig:iiwa_cost_cbundles}.
This clearly demonstrates that the lower cost achieved by sampling is almost entirely due to the lack of restriction to a specific IK branch $\kappa=\kappa_0$.
If we run the new formulation using the IK branch with the best solution, $\kappa^*$, then it is often able to find the globally optimal solution.
Although the old formulation is not restricted to a specific branch, it is not able to consistently find the branch of the optimal solution, and $34\%$ of the time, it produces a solution on the same branch as the initial guess.

Restricting to $\kappa_0$ in the new formulation may make the problem infeasible, since the set of feasible solutions in configuration space could potentially not contain any part of the manifold specified by $\kappa_0$. We explore this effect by choosing a random pose and a random $\kappa_0$ and seeing whether dense sampling can locate a solution. In $94\%$ of cases it can, which estimates a limit our success rates in the new formulation in \cref{tab:success_rates} to $94\%$ for the arm on a table.

\subsection{Grasp Selection IK}
\label{sec:experiments:grasp_selection}

Our next experiment expands the IK problem to target a range of poses that grasp an object.
The setup is equivalent to that of \Cref{fig:experimental_setup:arm_on_a_table}, with an array of bins and shelves on a table, except the gripper has longer fingers in order to grasp the mug.

To allow the optimizer to select a grasp, we require that a specified point in between the fingers be coincident with the central axis of a mug.
Unlike the grasp selection problem studied in~\cite{dai2019global}, we do not constrain the orientation of the gripper.
Instead, we rely on the collision-free constraint to ensure we select a reasonable grasp.
This constraint is linear in terms of the decision variables of the new formulation, but nonlinear in terms of the joint angles.
We selected 40 random targets, and ran the nonlinear solvers on 40 random initial guesses per target.
See \Cref{appx:experiment_implementation:grasp_selection} for further details.

\begin{table*}
    \renewcommand{\arraystretch}{1.25}
    \setlength{\tabcolsep}{3pt}
    \centering
    \begin{tabularx}{\textwidth}{|c|VC|C|VC|C|VC|C|Vc|>{\centering\arraybackslash}p{1cm}|V>{\centering\arraybackslash}p{1cm}|} \hline
        \multirow{3}{*}{Experiment} &
        \multicolumn{2}{c|V}{\multirow{2}{*}{\parbox{2.8cm}{\centering Interior Point\\(IPOPT)}}} &
        \multicolumn{2}{c|V}{\multirow{2}{*}{\parbox{2.8cm}{\centering Augmented Lagrangian\\(NLOPT)}}} &
        \multicolumn{2}{c|V}{\multirow{2}{*}{\parbox{2.8cm}{\centering Sequential Quadratic\\Programming (SNOPT)}}} &
        \multicolumn{2}{c|V}{\multirow{2}{*}{\parbox{2.0cm}{\centering Global-IK}}} &
        \multirow{3}{*}{Sampling}
        \\

        & \multicolumn{2}{c|V}{} & \multicolumn{2}{c|V}{} & \multicolumn{2}{c|V}{} & \multicolumn{2}{c|V}{} &
        \\ \cline{2-9}
        & New & Old & New & Old & New & Old & Fast & Precise &
            \\ \hline
        Arm on a Table & \bf 14.1 & 14.6 & 14.3 & \bf 11.3 & \bf 14.2 & 15.0 & 5.5 \goldstar & 5.8 & 6.1
            \\ \hline
        Grasp Selection & 9.22 & \bf 5.83 & 10.9 & \bf 3.6 \goldstar & \bf 10.9 & 14.0 & \tableNA & \tableNA & 7.44
            \\ \hline
        Mobile Manipulator & 8.46 & \bf 7.28 & 8.62 & \bf 3.44 \goldstar & 10.8 & \bf 6.81 & \tableNA & \tableNA & 14.58
            \\ \hline
        Bimanual Mobile Manipulator & 35.2 & \bf 25.5 \goldstar & \bf 30.6 & \tableNA & 37.7 & \bf 29.8 & \tableNA & \tableNA & 43.8
            \\ \hline
    \end{tabularx}
    \caption{
        Average optimal costs for each experiment, restricted to successes.
        For each solver, the formulation with the better optimal cost is bolded, and the overall lowest cost has a gold star.
        We do not include the humanoid stability experiments, as these were pure feasibility problems.
        The old formulation often finds lower cost solutions, in part because it is not restricted to a single branch of the IK function.
        (See \Cref{fig:iiwa_cost_cbundles} for further discussion.)
    }
    \label{tab:optimal_costs}
\end{table*}

\begin{table*}
    \renewcommand{\arraystretch}{1.25}
    \setlength{\tabcolsep}{1.25pt}
    \centering
    \begin{tabularx}{\textwidth}{|c|VC|C|VC|C|VC|C|V>{\centering\arraybackslash}p{1cm}|>{\centering\arraybackslash}p{1cm}|Vc|} \hline
        \multirow{3}{*}{Experiment} &
        \multicolumn{2}{c|V}{\multirow{2}{*}{\parbox{2.8cm}{\centering Interior Point\\(IPOPT)}}} &
        \multicolumn{2}{c|V}{\multirow{2}{*}{\parbox{2.8cm}{\centering Augmented Lagrangian\\(NLOPT)}}} &
        \multicolumn{2}{c|V}{\multirow{2}{*}{\parbox{2.8cm}{\centering Sequential Quadratic\\Programming (SNOPT)}}} &
        \multicolumn{2}{c|V}{\multirow{2}{*}{\parbox{2.0cm}{\centering Global-IK}}} &
        \multirow{3}{*}{Sampling}
        \\

        & \multicolumn{2}{c|V}{} & \multicolumn{2}{c|V}{} & \multicolumn{2}{c|V}{} & \multicolumn{2}{c|V}{} &
        \\ \cline{2-9}
        & New & Old & New & Old & New & Old & Fast & Precise &
            \\ \hline
        Arm on a Table & 0.14 (0.11) & \bf 0.07 (0.04) & 0.98 (0.84) & \bf 0.24 (0.16) & 0.025 (0.025) & \bf 0.02 (0.02) \goldstar& 289 & 3362 & 0.101
            \\ \hline
        Grasp Selection & 0.45 (0.46) & \bf 0.15 (0.15) \goldstar & 2.68 (3.84) & \bf 0.81 (1.13) & 1.07 (1.47) & \bf 0.36 (0.26) & \tableNA & \tableNA & 1.69
            \\ \hline
        Mobile Manipulator & 0.68 (0.52) & \bf 0.17 (0.12) \goldstar & 1.37 (3.52) & \bf 0.23 (0.32) & 0.90 (1.17) & \bf 0.18 (0.17) & \tableNA & \tableNA & 0.60
            \\ \hline
        Bimanual Mobile Manipulator & 1.33 (0.64) & \bf 0.40 (0.16) & \bf 6.01 (4.76) & \tableNA & 2.25 (2.30) & \bf 0.18 (0.09) \goldstar & \tableNA & \tableNA & 1.73
            \\ \hline
        Humanoid Stability (Inequality) & \bf 1.26 (1.12) \goldstar & 3.45 (1.82) & \bf 25.82 (17.82) & \tableNA & \tableNA & \tableNA & \tableNA & \tableNA & \tableNA
            \\ \hline
        Humanoid Stability (Equality) & \bf 1.61 (1.31) \goldstar & 1.65 (1.50) & \bf 36.37 (24.00) & \tableNA & \tableNA & \tableNA & \tableNA & \tableNA & \tableNA
            \\ \hline
    \end{tabularx}
    \caption{
        Mean runtimes for each experiment.
        For each solver, the formulation with the shortest runtime is bolded, and the overall shortest runtime has a gold star.
        For the nonlinear optimizers, the entry in parentheses is restricted to only the successes.
        (This is not done for Global-IK, since we do not set a timeout, and not done for sampling, since its runtime is constant.)
    }
    \label{tab:runtimes}
\end{table*}

Unsurprisingly, expanding the set of feasible end-effector poses greatly improves the performance of the old formulation.
The new formulation still has a high success rate, albeit with a slightly worse cost performance.
Part of this can be again attributed to the restriction $\kappa=\kappa_0$ in the new formulation.
If we compare solutions from the old and new formulation that are from the same branch of the analytic IK function, the costs are extremely similar, as in the previous experiment.

On the other hand, the larger solution space greatly reduces the success of sampling as a baseline. 
In five dimensions, when we budget the sampling runtime to roughly match the optimization methods (8192 total samples), the success rate falls dramatically.
This is likely due to the set of feasible orientations being a small subset of the total sample space. Thus, it was more difficult for sampling to find a feasible orientation to hold the mug. 

However, the new formulation also suffers from the non-convexity of the cost function with respect to the orientation within the decision variable $\monogram{o}{W}{G}$, making the new formulation much more susceptible to local minima.
Minimizing the cost, rather than finding a feasible solution, is far more difficult in this setup.
When solving a pure feasibility problem, both formulations achieve a $100\%$ success rate, even with the challenges associated with the collision-avoidance constraint.

While sampling cannot find a local minimum, we can instead compute the minimum cost found by the old and new formulation of a single mug over all 40 initial guesses. Since in our experiment, the nominal cost formula is independent of initial guess, comparing minimal results over all initial guesses helps us understand if both formulations are able to locate a local minimum. Indeed, taking the minimal cost over 40 initial guesses for the old formulation and new formulation on a single target, we find that they often find the exact same minimum, which indicates that both formulations, allowing for multiple initial guesses, likely found the global minimum cost. 

\subsection{A Basic Mobile Manipulator}
\label{sec:experiments:kmr_iiwa}

In the following experiment, a KMR iiwa is operating around a central table with shelves.
Its end-effector is a Schunk WSG-50, as shown in \Cref{fig:experimental_setup:kmr_iiwa}.
We return to targeting a specific gripper pose, and solve for a very similar optimization problem to the one in \Cref{sec:experiments:arm_on_a_table}: we impose a constraint on the end effector pose $\monogram{X}{W}{G} = \monogram{X}{W}{G}_{\des}$, as well as joint limits and collision-avoidance.
We selected 100 end-effector poses within the shelves, and ran the nonlinear solvers on 100 random collision-free initial guesses per target.
See \Cref{appx:experiment_implementation:kmr_iiwa} for details.

We again find that the new formulation is more robust in terms of success rate but achieves a worse cost than the old formulation, presumably due to the simpler cost landscape.
In addition, the old formulation performs much better for the mobile base then for the arm on the table.
This is likely since the base can adjust for small errors of horizontal position and yaw more directly.
As with the grasp selection experiment, removing the cost results in success rates of $95\%$ and $98\%$ for the old and new formulations, respectively.
This matches the conventional wisdom that IK is easier when the robot has a mobile base~\cite[\S 7.6]{russtedrake2024manipulation}.

Our only additional constraint to the new formulation is the reachability constraint \eqref{eq:new_formulation:reachable}.
Although the clipping operation within the IK function loses gradient information, the gradients from the probing functions provide useful information to move the robot towards the target.
In many instances, the optimizer was able to move the base such that the target was reachable within just a few iterations.

\subsection{A Bimanual Mobile Manipulator}
\label{sec:experiments:pr2}

Next, we implement an experiment with the Willow Garage PR2, a bimanual mobile manipulator.
Above its mobile base, the PR2 consists of a prismatic torso lift joint and two 7DoF arms.
We use an improved version of the analytic IK solution of Ramezani and Williams~\cite{ramezani2015pr2} -- our improvements are detailed in \Cref{appx:ik_implementation:pr2}.
The experimental setup is similar to the previous experiment, and is shown in \Cref{fig:experimental_setup:pr2}.
We now can set two targets, one for the left hand, one for the right hand, resulting in two constraints.
Each constraint corresponds to a separate kinematic chain, with only the base pose and torso lift coupling them.
For the new formulation, the decision variables are the left and right end-effector poses, the respective self-motion parameters, the base pose, and the torso lift.
We now have two reachability constraints in the new formulation, and as in \Cref{sec:experiments:kmr_iiwa}, we do not encode hard constraints into the problem with log-barriers.
We selected 40 random targets, and ran the nonlinear solvers on 40 random initial guesses per target.
See \Cref{appx:experiment_implementation:pr2} for further details.

The two equality constraints present a significant challenge for the optimizer. 
We observe a new failure mode when the robot cannot figure out the correct base orientation, electing to cross its arms and and ultimately getting stuck in an infeasible local minimum instead of spinning around so that both arms are reachable.
Hence, the success was highly dependent on whether the initial guess had the base in a good orientation.
At the same time, for some sets of targets, such as those shown in \Cref{fig:experimental_setup:pr2}, we find that the new formulation is better for collision-avoidance while crossing arms, locking the end-effector poses into place and allowing the base and first joint to move around to avoid collision.

Of the solvers we tried, NLOPT always resulted in infeasibility for the old formulation, even with extremely low feasibility tolerances. 
Overall, we find that the new formulation is much more robust in success rates, while the old formulation is faster and can find better optimal costs when it is successful.

\subsection{Humanoid Stability}
\label{sec:experiments:hubo}

\begin{figure*}
    \centering
    \begin{subfigure}[b]{0.245\linewidth}
        \centering
        \includegraphics[width=\linewidth]{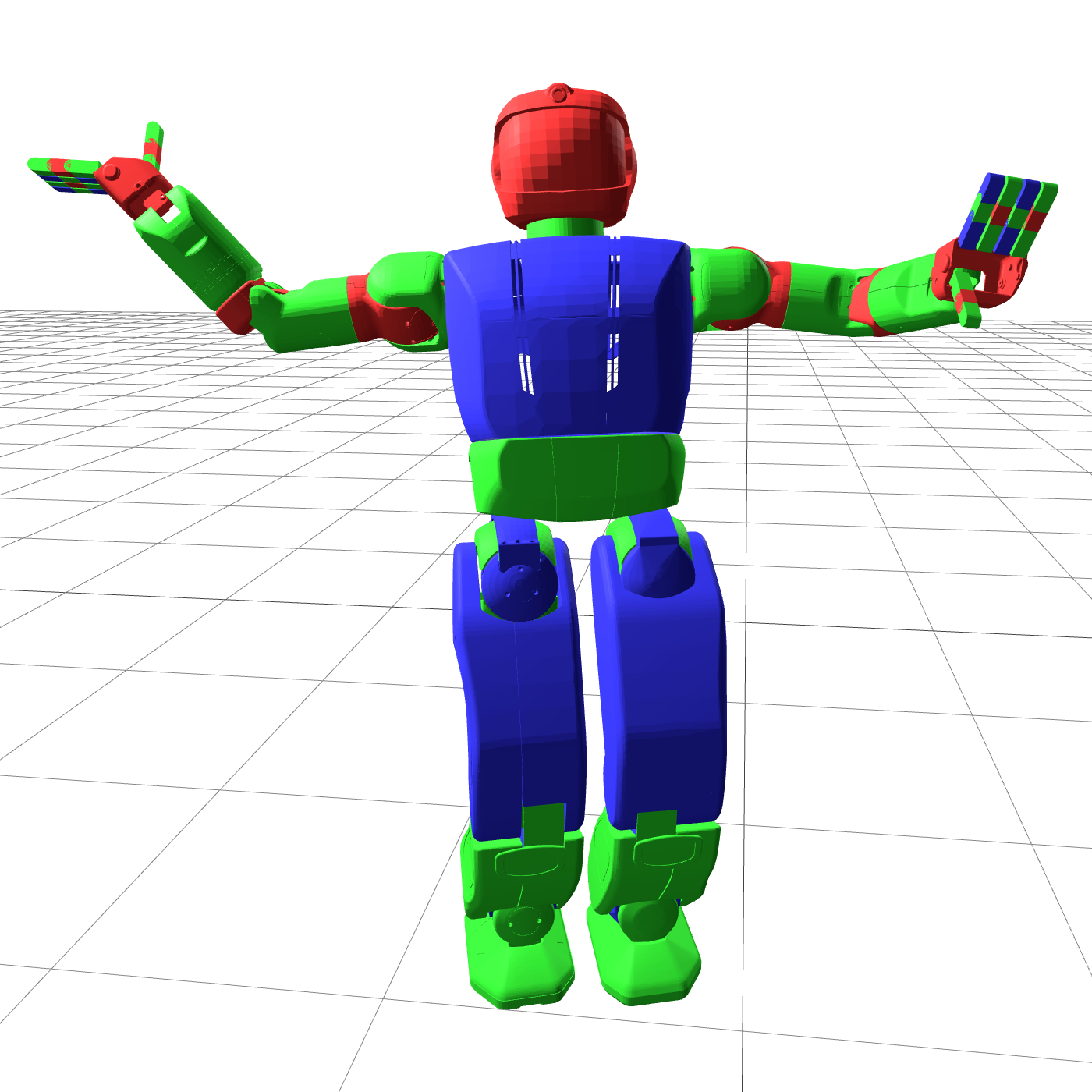}
        \caption{Old formulation, equality.}
        \label{fig:hubo_diverse_solutions:old_equality}
    \end{subfigure}
    \begin{subfigure}[b]{0.245\linewidth}
        \centering
        \includegraphics[width=\linewidth]{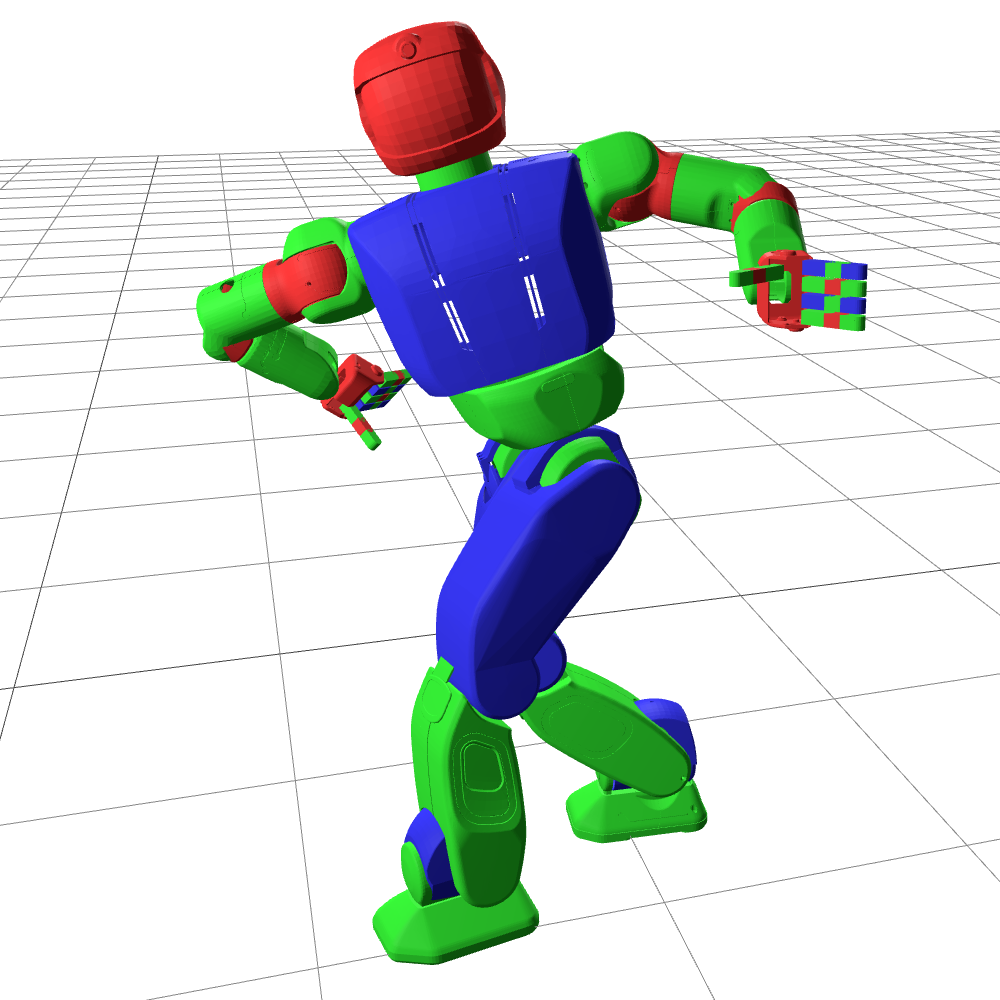}
        \caption{Old formulation, inequality.}
        \label{fig:hubo_diverse_solutions:old_inequality}
    \end{subfigure}
    \begin{subfigure}[b]{0.245\linewidth}
        \centering
        \includegraphics[width=\linewidth]{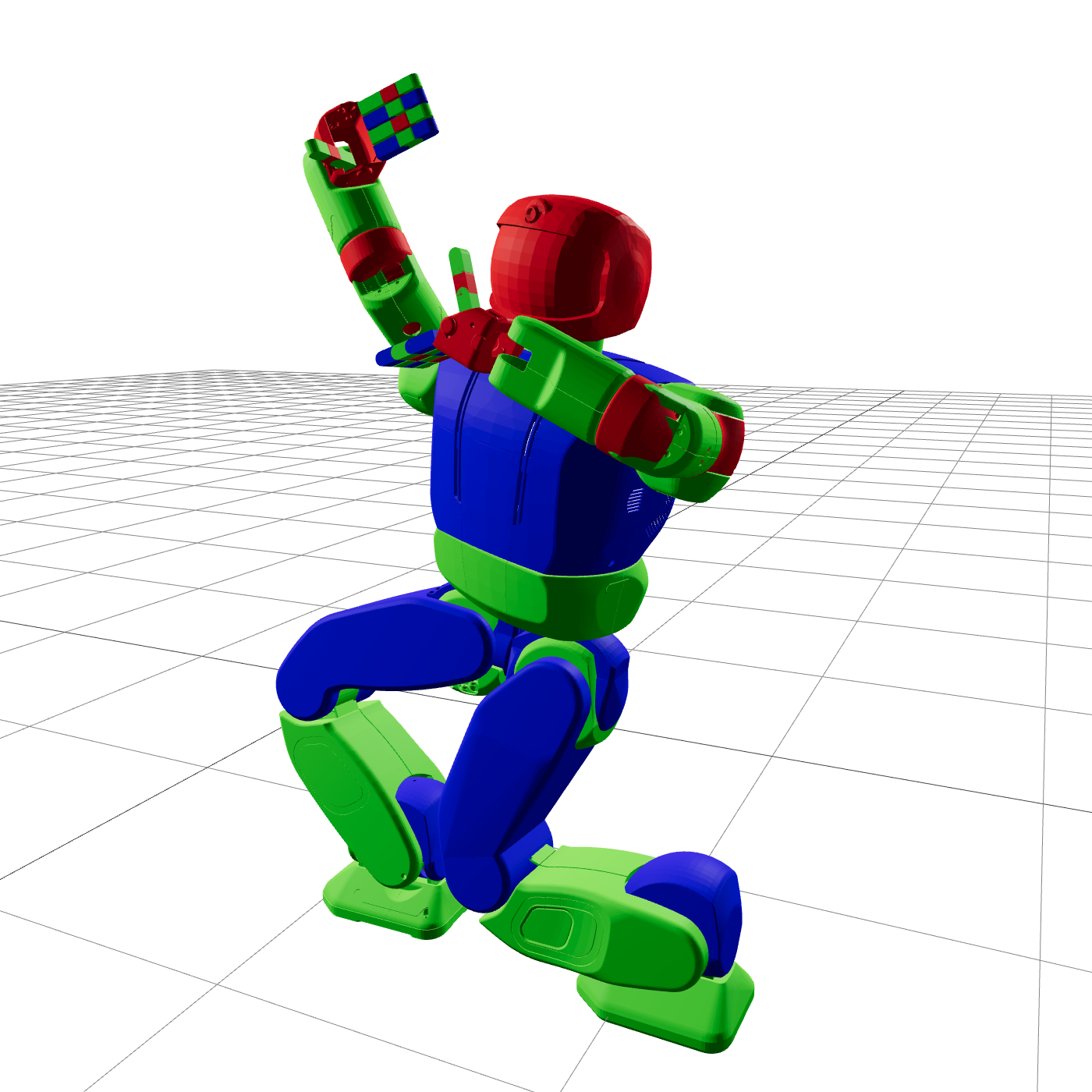}
        \caption{New formulation, equality.}
        \label{fig:hubo_diverse_solutions:new_equality}
    \end{subfigure}
    \begin{subfigure}[b]{0.245\linewidth}
        \centering
        \includegraphics[width=\linewidth]{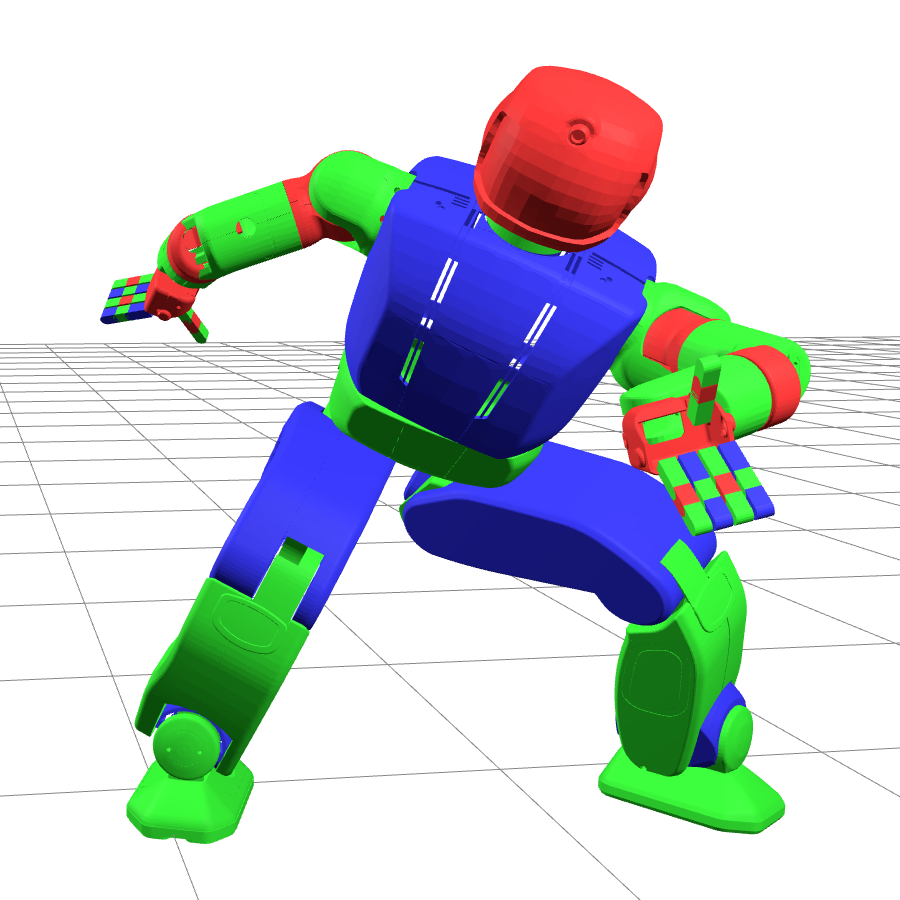}
        \caption{New formulation, inequality.}
        \label{fig:hubo_diverse_solutions:new_inequality}
    \end{subfigure}
    \caption{
        Diverse solutions from the Hubo experiments, from the different formulations and stability constraint representations. All solutions were obtained with IPOPT. 
    }
    \label{fig:hubo_diverse_solutions}
\end{figure*}

In this experiment, we consider a humanoid reaching task, where the robot must reach to a given target pose, while simultaneously placing its feet so as to guarantee stability.
If both feet are flat on the ground, static stability can be ensured by keeping the center of mass above the \emph{support polytope}: the convex hull of the points on the foot making contact with the ground.
These quantities can all be computed efficiently as functions of the robot's configuration, making them amenable for optimization.

We use the Hubo 2+, henceforth just called ``Hubo'', a humanoid robot with two 6DoF arms, two 6DoF legs, and a revolute waist.
Hubo also has a 3DoF neck, which we treat as a free variable but otherwise ignore, and it has additional degrees of freedom in its hands, which we treat as fixed.
Thus, including the 6DoF pose of its torso, we have $d=34$ degrees of freedom.
In our simulations, we have modified the robot description so that its arms' kinematics match those described in the analytic IK papers~\cite{ali2010hubo,park2012hubo,oflaherty2013hubo}; see \Cref{appx:ik_implementation:hubo} for further discussion.

We seek to find a stable configuration where the right hand achieves a given target pose.
Stability is enforced by requiring both feet be flat on the ground, with the center of mass over the support polytope.
We also require the robot satisfy joint limits and be collision-free.
We do not add a cost, since this frequently causes the optimizer to sacrifice feasibility in order to find a lower objective value, and ultimately get stuck at an infeasible local minimum.

In the new formulation, all equality constraints are linear, except for the constraint enforcing that the center of mass be over the support polytope.
This allows us to study the performance of the new formulation in the presence of additional nonlinear equality constraints.
However, the set of feasible configurations is still positive measure, and we derive an inequality formulation of the stability constraint.
We selected 100 random targets.
All details are contained in \Cref{appx:experiment_implementation:hubo}.

In the new formulation, we do include log barrier costs to enforce joint limits and reachability.
Within these costs (and anytime the analytic IK is called), NLOPT required that the inputs to $\arccos$ be clipped to $[-1+\epsilon,1-\epsilon]$ and inputs to $\log$ be clipped to $[\epsilon,\infty)$.
In these experiments, we choose $\epsilon=10^{-6}$.
Note that such clipping does \emph{not} work with IPOPT, leading to high failure rates.

Of all of the solvers we tried, only IPOPT was usable.
SNOPT immediately returns infeasible for both formulations.
NLOPT immediately returns infeasible for the old formulation, but it was sometimes able to solve the new formulation (albeit at a much lower success rate and higher runtime).
It is unclear how to represent the stability constraints as part of the Global-IK framework, and sampling is intractable -- the kinematic constraints only remove $12$ degrees of freedom, leaving a $22$ dimensional space with a feasible set of extremely small measure.

In both the new and old formulation, the optimizers were very sensitive to the initial guess (with random guesses yielding low success rates).
This is in part due to the complexity of the optimization problem, but also due to the stability constraint, which is only meaningful as long as the feet are actually flat on the ground.
To avoid this problem, we use a nominal stable configuration as an initial guess for both formulations.
Note that the optimizer still discovers diverse solutions to the problem, with example solutions found by the new and old formulation shown in \Cref{fig:hubo_diverse_solutions}.
Finally, for the old formulation, we needed to set a relatively loose constraint satisfaction tolerance of $10^{-4}$.

With the formulation of the stability constraint as a nonlinear equality constraint, both the old and new formulation achieve a high success rate and fast runtimes using IPOPT.
This strongly supports our claim that our new formulation is still effective in the presence of additional nonlinear equality constraints.
And in the case of the inequality formulation of the stability constraint, in which the new formulation has a relative interior, the new formulation achieves an even faster runtime without a decrease in success rate.

\subsection{Scaling with the Number of Redundant Degrees of Freedom}
\label{sec:experiments:scaling}

One major question of theoretical interest is how the two formulations compare as the degree of kinematic redundancy increases.
Although the previous experiments span a wide range of self-motion manifold dimension, the variety of other factors that differ between experiments make it hard to isolate the cause of any trends.
Thus, we consider two experimental setups where we can incrementally add joints to a robot arm while minimizing changes to any other aspects.

\subsubsection{2D Setup}
\label{sec:experiments:scaling:2d}

In this experiment, we create a 2-dimensional link-chain with $n$ links, each with length $1/n$, for different values of $n$.
Each pair of links is connected with a revolute joint.
We fix the end-effector target, while imposing a quadratic joint centering cost.
In the new formulation, the joint angles of all but the last three joints are used as the self-motion parameters, with a standard 3R analytic solution being used to compute the remaining three joint angles.
While not including key constraints like collision avoidance, this provides a level of abstraction for testing the scaling of the new formulation as the degree of self-motion, i.e the number of joints, increases.
For each number of links, we selected 300 random targets, and ran the nonlinear solvers with one random initial guess per target.
All details are contained in \Cref{appx:experiment_implementation:scaling_2d}.

With and without the joint centering cost, we find that the new formulation performs best at low degrees of self-motion, with the median iterations per solve gradually increasing. At high dimensions the old formulation performs better.
Even setting the old formulation's optimality tolerance very low, the new formulation is able to find a better cost in the 2d setup.
This is likely because the inverse kinematics function for the last three joint angles is tractable enough to calculate useful gradients for the optimizer.

\begin{figure*}
    \centering
    \begin{subfigure}[b]{0.49\linewidth}
        \centering
        \includegraphics[width=0.9\linewidth]{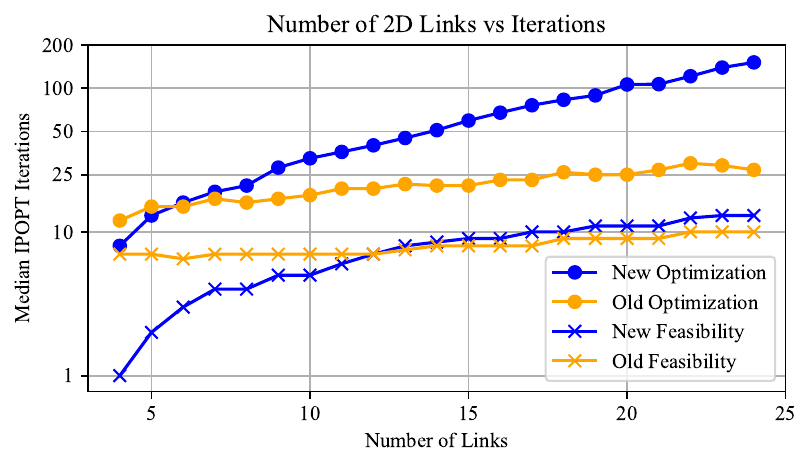}
        \caption{Median iterations plotted against number of joints (log-scale). }
        \label{fig:2d_scaling:iterations}
    \end{subfigure}
    \begin{subfigure}[b]{0.49\linewidth}
        \centering
        \includegraphics[width=0.9\linewidth]{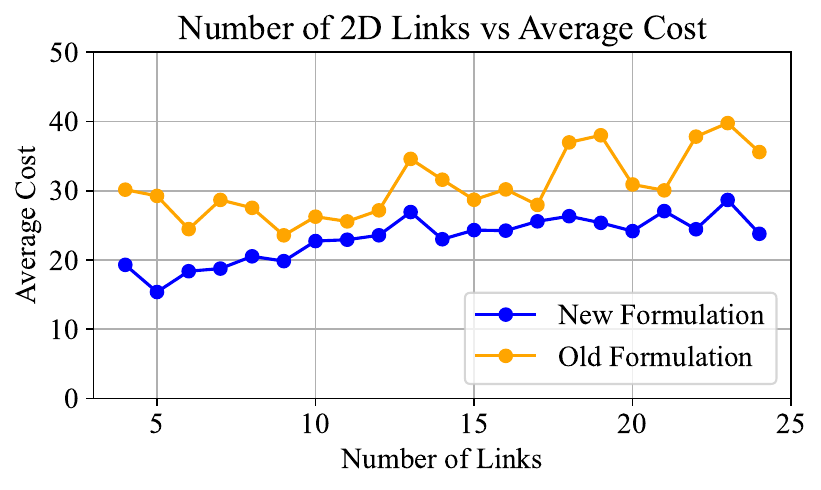}
        \caption{Average cost plotted against number of joints.}
        \label{fig:2d_scaling:cost}
    \end{subfigure}
    \caption{Results for old and new formulation of the 2-dimensional link-chain IK.}
    \label{fig:2d_scaling}
\end{figure*}

\subsubsection{3D Setup}
\label{sec:experiments:scaling:3d}

We expand our scaling set-up to a 3d arm.
We craft an $n+7$ link arm, connected by revolute joints.
The last $7$ joints have a sphere-revolute-sphere kinematic structure that matches the KUKA iiwa, so we can leverage that analytic IK solution.
Once again, we consider a quadratic joint centering cost, and we also solve the analogous feasibility problem, since we know that one disadvantage of the new formulation is that the cost is far more complicated.
For each number of links, we selected 200 random targets, and ran the nonlinear solvers with one random initial guess per target.
All details are contained in \Cref{appx:experiment_implementation:scaling_3d}.

\begin{figure*}
    \centering
    \begin{subfigure}[b]{0.49\linewidth}
        \centering
        \includegraphics[width=0.9\linewidth]{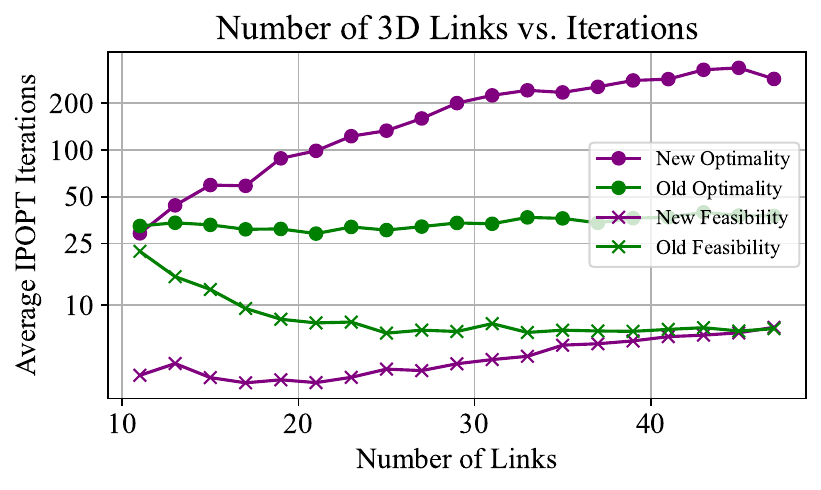}
        \caption{Average iterations plotted against number of joints (log-scale).}
        \label{fig:3d_scaling:iterations}
    \end{subfigure}
    \begin{subfigure}[b]{0.49\linewidth}
        \centering
        \includegraphics[width=0.9\linewidth]{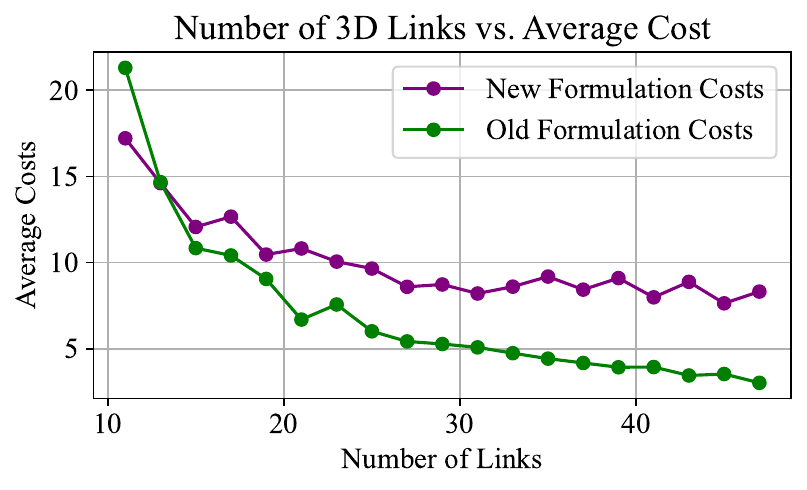}
        \caption{Average cost plotted against number of joints.}
        \label{fig:3d_scaling:cost}
    \end{subfigure}
    \caption{Results for old and new formulation of the 3-dimensional link-chain IK.}
    \label{fig:3d_scaling}
\end{figure*}

For arms with 15 or more joints, both the old and new formulation are able to achieve a 100\% success rate in the optimality problem.
For arms under 15 joints, the new formulation maintains its $100\%$ success rate, while the old formulation drops to around $90\%$ for 11 link arms. 
As expected, the old formulation is able to find a better optimal cost then the new formulation.
We plot the behavior in \Cref{fig:3d_scaling:cost}.

We also plot the number of iterations needed to converge to optimality against the number of joints (not including solve failures).
We set the tolerances at a moderately low value, at $10^0$, as the performance of the new formulation drastically deteriorated with lower optimality tolerances.
Additionally, we set a timeout of 30 seconds, and if a timeout occurs, we capture the minimum cost achieved which satisfied all constraints.

We find that the performance of the new formulation is best with a low number of joints, with the number of iterations growing dramatically with the number of joints.
The cost landscape of the new formulation is much worse than the old, and it takes significantly more iterations for the new formulation to find close to an optimal cost.
The old formulation performs equivalently to the new in terms of number of iterations when there is little kinematic redundancy, but the number of iterations needed remains relatively constant as the number of joints increases.

Without a cost, the number of iterations required to find a feasible solution is consistently very low for the new formulation, only growing slightly with the number of joints.
The number of iterations required for the old formulation is initially larger than the old formulation, but decreases as a higher degree of self-motion is possible, matching the new formulation past $40$ joints.

\section{Discussion}
\label{sec:discussion}

In this paper, we have presented a new formulation for optimization-based inverse kinematics, that leverages a known analytic solution for the robot as a change of variables.
This allows us to write IK optimization problems with the end-effector and self-motion parameters as decision variables, eliminating nonlinear equality constraints.
We compare our approach with the corresponding standard formulation across three solvers, each representing a different broad class of nonlinear optimization algorithms, and we also compare to two baseline algorithms.
Diverse practical experiments include shelf reaching for fixed and mobile robots, grasp selection, and humanoid stability, demonstrating the relevancy of our contributions to realistic robotics tasks under consideration today.

To obtain these results, we have also presented improvements to existing analytic IK solutions for the PR2 mobile manipulator and the Hubo2+ humanoid.
And to strengthen the Global-IK baseline~\cite{dai2019global}, we find a convex cost function that better approximates the desired quadratic joint-centering cost, and explore combining this approach with modern free-space decomposition algorithms to solve collision-free IK problems without user intervention.

When comparing the two formulations, the new formulation consistently achieved equal or higher success rates.
Results were mixed as to which formulation is faster, and a fair comparison would require that all optimization problems be written in the same language.
(Almost all of the old formulation is in C++, but many of the calculations for the new formulation are implemented Python.)
Finally, the new formulation is consistently able to obtain extremely feasibility tolerances (on the order of $10^{-17}$ or better), many orders of magnitude below what solvers can normally achieve in the presence of nonlinear equality constraints.
The poorer performance in optimal cost remains a significant limitation of the new formulation.
But our results strongly suggest that this new formulation is useful in practice, particularly when additional challenging constraints are present, and current success rates are low.

The applicability of our method is aided by the fact that many of the most popular robot arms today have analytic IK solutions.
We have leveraged hand-crafted solutions to ensure compatibility with Drake's automatic differentiation, whereas automatically generated IK solutions (such as those from IKFast~\cite{diankov2010ikfast} or IK-Geo~\cite{elias2025ikgeo}) would have to be modified for compatibility.
But this is an engineering issue, rather than a fundamental roadblock, and these automated approaches could be upgraded to generate compatible code, or otherwise support computing derivatives.

As for the remaining robot arms that do not have known analytic IK solutions, one could potentially use other numerical or learning-based IK approaches, or even the numerical algebraic geometry methods for IK, together with this framework, so long as it is possible to differentiate through them as a mapping.
Whether such strategies are algorithmically practical or numerically tractable remains an open question worthy of further study.

\section{Acknowledgements}
\label{sec:acknowledgements}
The authors thank Hongkai Dai (Toyota Research Institute) and Seiji Shaw (Massachusetts Institute of Technology) for their many helpful comments and suggestions.
This project was supported by Amazon.com, PO No. 2D-15693043 and 2D-15694085, the National Science Foundation under Grant No. DMS-2022448, the National Science Foundation Graduate Research Fellowship Program under Grant No. 2141064, and the Robotics and AI Foundation.
Any opinions, findings, and conclusions or recommendations expressed in this material are those of the author(s) and do not necessarily reflect the views of the National Science Foundation.
They also do not necessarily reflect the views of the other sponsors acknowledged in this work.

\bibliographystyle{IEEEtran}
\bibliography{ref.bib}

\appendices

\crefalias{section}{appendix}
\crefalias{subsection}{appendix}

\renewcommand{\thesubsectiondis}{\Alph{section}.\arabic{subsection}}
\renewcommand{\thesubsection}{\Alph{section}.\arabic{subsection}}

\section{Implementation of the Analytic IK Mappings}
\label{appx:ik_implementation}
In this appendix, we discuss implementation details of the analytic IK mappings used in this paper.
While computer-generated solutions, such as \cite{diankov2010ikfast}, are well-developed, we choose not to use computer-generated solutions for two reasons:
\begin{enumerate}
    \item We need to be able to differentiate through the mappings.
    \item We need to extract intermediate values before they are passed into domain-restricted functions like $\arccos$.
\end{enumerate}
To the authors' knowledge, no existing method for automatically generating analytic IK solutions satisfies both desiderata.
Due to the challenge of modifying highly-optimized computer-generated solutions, we produced our own implementations of the required analytic IK functions.
Our solutions are written in python, and carefully designed to be compatible with Drake's automatic differentiation capabilities.%

The analytic IK solution for the KUKA iiwa presented by Faria et al.~\cite{faria2018iiwa} required no significant modifications.
We simply added clipping for the input to the $\arccos$ function, which was used in Eqs. (18) and (23), along with functionality to return the unclipped values.
This analytic IK solutions was used for \Cref{sec:experiments:arm_on_a_table,sec:experiments:grasp_selection,sec:experiments:kmr_iiwa}, as the KMR iiwa is just an iiwa arm mounted on a mobile base.

\subsection{PR2 Analytic IK}
\label{appx:ik_implementation:pr2}

The implementation of the analytic inverse kinematics for the PR2 was based on \cite{ramezani2015pr2}, where the first joint angle is chosen as the redundancy parameter.
The methodology of this paper produces both true and extraneous solutions, which must be filtered out by checking each possible solution against the forward kinematic map.
Instead, we further specify the solutions by solving for both $\sin \theta_i$ and $\cos \theta_i$ such that $\theta_i \in [0, 2\pi]$ is uniquely determined with $\arctan$. 
We find a set of eight closed-form solutions, which we index as $\kappa = \{g_i\} \in \{ -1, 1\}^3$ to match the convention used for the iiwa IK.
We reproduce the relevant equations below.

Label the joint angles $\theta_0, \theta_1, \ldots, \theta_6$, with corresponding DH parameters for link $i$ being $\alpha_i, d_i, a_i$. For simplicity, let
\begin{equation*}
    s_i := \sin\theta_i, \;\;\; s_{ij} := \sin\theta_i\sin\theta_j, \;\;\; s_{ijk} := \sin\theta_i\sin\theta_j\sin\theta_k,
\end{equation*}
with analogous notation for cosine.

First, given a desired end-effector pose $(\mathbf{p_d, o_d})$, we clip and solve for $\cos \theta_3$ as in \cite{ramezani2015pr2}, and parameterize the sign of $\theta_3$ with $g_0$. 
We then solve for $\theta_2$ by taking Eq 15. in \cite{ramezani2015pr2}, to get \begin{equation}
    d_4s_{32} = p_yc_0-p_xs_0,
\end{equation}
where $\mathbf{p_d} = (p_x, p_y, p_z)$, solving and clipping the solution for for $\sin \theta_2$, and parameterizing the two solutions with $g_1$. 
This differs from the original calculation of $\theta_2$, which solves for $s_2^2$ and leads to two extraneous solutions.

Following this, \cite{ramezani2015pr2} defines variables 
\begin{gather}
    u_1 = a_0+u_3c_1+u_4s_1\\
    u_2=s_{32}d_4\\
    u_3=c_2s_3d_4 \\
    u_4=d_2+c_3d_4,
\end{gather}
and shows that \begin{equation}
    \cos (\theta_1 + \vartheta) = \frac{p_z}{\sqrt{u_3^2+u_4^2}},
\end{equation}
where $\vartheta = \tan^{-1}\frac{u_3}{u_4}$. 

We further find \begin{gather}
    \sin (\theta_1 + \vartheta) = \frac{u_1 - a_0}{\sqrt{u_3^2+u_4^2}} \\
    u_1 = \frac{p_x + u_2s_0}{c_0},
\end{gather}
allowing us to uniquely specify $\theta_1$. 

To determine the last few joint angles, we solve for \begin{equation}
    ^3\mathbf{o}^6 = {}^3\mathbf{o}^0\mathbf{o_d},
\end{equation}
where the left hand side is entirely a determined by the $\theta_0, \ldots \theta_3$, while $(\theta_4, \theta_5, \theta_6)$ has two solutions determined by 
\begin{equation}
    ^3\mathbf{o}^6 = \begin{pmatrix}
        c_{456}-s_{46} & -c_{45}s_6-s_4c_6 & c_4s_5\\
        s_4c_{56}+c_4s_6 & -s_{46}c_5 + c_{46} & s_{45} \\ 
        -s_5c_6 & s_{56} & c_5
    \end{pmatrix}.
\end{equation}
These two solutions are parameterized by $g_2$. Since the PR2 wrist is a spherical wrist, the transformation between the two solutions are well-understood, or can be derived as: 
\begin{equation}
    (\theta_4, \theta_5, \theta_6) \Leftrightarrow (\theta_4+\pi, -\theta_5, \theta_6 + \pi).
\end{equation}

\subsection{Hubo Analytic IK}
\label{appx:ik_implementation:hubo}

The implementation of analytic IK for Hubo was based on \cite{ali2010hubo,park2012hubo,oflaherty2013hubo}.
First, we note that these papers derive IK solutions for an idealized kinematics model of the arm, where the axes of the third and fifth joint intersect with the axis of the fourth joint (the ``elbow'').
On the actual robot, the fourth joint is offset; this is visible, for instance, in Fig. 1 of \cite{oflaherty2013hubo}.
This leads to an error of up to several centimeters between the desired and realized end-effector poses from these analytic IK solutions, so for the purposes of this paper, we modify the URDF of Hubo to match the idealized kinematics.

In addition, \cite{oflaherty2013hubo} claims that there are errors in the calculations of \cite{ali2010hubo,park2012hubo} and presents a new derivation of the analytic IK solution.
However, this work did not actually identify where the mistakes are located.
Furthermore, the authors changed the DH parameters to describe the robot, making it difficult to identify the differences or combine the solutions.
Another issue is the presence of branching logic (e.g. the conditional use of the ``$\operatorname{wrapToPi}$'' routine in \cite{oflaherty2013hubo}).
In practice, we found that slightly modifying \cite{oflaherty2013hubo} arrived us to a true closed form solution.

In multiple instances, both \cite{park2012hubo} and \cite{oflaherty2013hubo} branch out a case \begin{gather}
    x = a\sin\theta_i\\
    y = a\cos \theta_i
\end{gather}
such as after (40) or (50) in \cite{oflaherty2013hubo}, where the authors inconsistently include an expression of the form $\theta_i = \theta_i + \pi$ when $a < 0$.
While making the numerics a little more complicated, we instead opt for $\theta_i = \operatorname{arctan2}(ay,ax)$ to remove the branching logic, with $a=0$ presenting singularity.
Note that $\operatorname{arctan2}(ay,ax)=\operatorname{arctan2}(y/a,x/a)$ for nonzero $a$.

\section{Implementation of the Global MICP IK Baseline}
\label{appx:micp_ik_implementation}

In \Cref{sec:experiments} we compare our performance to a mixed-integer convex optimization approach, Global-IK \cite{dai2019global}. 
Here, we first review the Global-IK method, including how collision avoidance is handled.
We then discuss our pipeline that combines Global-IK with the Clique Cover method for generating collision-free polytopes to cover free space.
Finally, we detail how we selected the ``fast'' and ``precise'' settings in our experiments.

Global-IK utilizes a mixed-integer convex relaxation of non-convex $\SO(3)$ rotation constraints. 
For instance, a rotation matrix $\monogram{o}{i-1}{i} = [\mathbf{u}_1 \quad\mathbf{u}_2 \quad \mathbf{u}_3]\in \mathbb{R}^{3\times 3}$, must satisfy non-convex bilinear quadratic constraints.
By partitioning the range of each variable into small chunks and taking linear outer-approximations, it is possible to model the problem as a mixed-integer convex program.
IK position and orientation constraints for the position of specific joints can then be readily formulated in terms of this convex relaxation as linear constraints:
\begin{gather}
    \monogram{p}{W}{i}
    = \monogram{p}{W}{(i-1)}
    + \monogram{o}{W}{(i-1)}\, \monogram{p}{\,(i-1)}{J} \\
    \monogram{o}{W}{(i-1)}\, \monogram{\hat{z}}{\,(i-1)}{i}
    = \monogram{o}{W}{i}\, \monogram{\hat{z}}{\,i}{i}
\end{gather}
with decision variables
$\monogram{o}{W}{i}, \monogram{o}{W}{i-1}, \monogram{p}{W}{i}, \monogram{p}{W}{i-1}$.
An implementation of Global-IK is included in Drake, via the \href{https://github.com/RobotLocomotion/drake/blob/v1.44.0/multibody/inverse_kinematics/global_inverse_kinematics.h}{\texttt{GlobalInverseKinematics}} class.

To impose collision avoidance constraints, Global-IK represents collision-free space as a union of polytopes and the collision geometry of the robot as a union of spheres. 
The collision-free constraint is then implemented as a mixed-integer convex linear constraint that each sphere is contained in a collision-free polytope \cite{dai2019global}.
Thus, the number of binary variables needed to encode this constraint is equal to the product of the number of spheres and polytopes.

Previously, collision-free polytopes for Global-IK in each scene were created by hand. 
In our pipeline we produce polytopes more generally through the Clique Cover method \cite{werner2024approximating}. The algorithm constructs a visibility graph with sampled collision-free points as vertices, and finds cliques within this graph. 
Afterwards, each clique is expanded into a large, full-dimensional polytope using an iterative region inflation (IRIS) algorithm \cite{petersen2023growing,werner2024faster}.
These algorithms are also implemented in Drake, and were originally meant to be used in configuration space.
We apply them in task space by modeling a dummy point robot, so that configuration space is equivalent to task space.
Thus, we can generate a set of collision-free polytopes in task space, stopping the process at a predetermined threshold of sampled scene coverage.

Increasing the desired coverage threshold comes with an increased number of polytopes.
This aggressively scales the number of binary variables in the mixed integer optimization problem, and thus the makes the problem much harder.
However, the polytope coverage that the polytopes forces an upper bound on the feasible joint configurations.
For example, the end-effector collision sphere must may entirely in at least one collision-free polytope for the problem to be feasible. 

We test each algorithm's performance by measuring how many polytopes are needed for differing levels of scene coverage over a variety of random seeds, and present the findings in \Cref{tab:iris_clique_cover}.
We also estimate how much of configuration space is covered by the polytopes, choose random collision-free poses, and check if all of their spheres are enveloped in at least one polytope. We collect data on a set-up is the similar to \Cref{fig:experimental_setup:arm_on_a_table}, except without mugs on the shelves.

\begin{table}[h]
    \renewcommand{\arraystretch}{1.25}
	\centering
	\begin{tabular}{|c|c|c|c|c|} \hline
        Iris & Coverage & Polytopes & Total Planes & C-space Covered\\ \hline
        \multirow{3}{*}{IrisNP}& 0.70 & 16.3 & 208 & 0.755\\ \cline{2-5}
        & 0.85 & 27.9 & 356 & 0.827\\ \cline{2-5}
        & 0.95 & 47.9 & 615& 0.881\\ \hline
        \multirow{3}{*}{IrisNP2} & 0.70 & 18.2 & 239 & 0.713 \\  \cline{2-5}
        & 0.85 & 33.5 & 443 & 0.810\\ \cline{2-5}
        & 0.95 & 54.7 & 771  & 0.884\\ \hline
        \multirow{3}{*}{IrisZO} & 0.70 & 19.1 & 364 & 0.831\\  \cline{2-5}
        & 0.85 & 32.2& 853 & 0.880\\ \cline{2-5}
        & 0.95 & 73.1 & 1935& 0.942\\ \hline
             
	\end{tabular}
	\caption{ 
        Average Performance of different Iris algorithms on creating collision-free polytopes.
	}         
	\label{tab:iris_clique_cover}
\end{table}

We found increasing the physical coverage of the collision-polytopes had diminishing returns on the gains in the configuration space covered. While the Iris from Clique Cover algorithm could find polytopes to cover small corners and spaces within the set-up, they were increasingly small or thin, often being unable to encircle a sphere of the collision-geometry of the iiwa.

At the same time, the resolution of relaxation does result in noticeable errors in the reconstructed end-effector pose, which makes the success rate of Global-IK in collision-free IK much lower than expected.
We specify two settings, fast and precise, in \Cref{tab:success_rates,tab:optimal_costs,tab:runtimes}. 
In the fast setting, the relaxation is loose, with larger discrepancies in position and orientation of up to ten centimeters (which can cause solutions to be found that are actually in collision), and we request 70\% scene coverage for our collision-free polytopes.
In the precise setting, the relaxation is stronger, with smaller inaccuracies, now on the scale of millimeters. We also request 85\% scene-coverage for our collision-free polytopes.

Since the decision variables of this problem are the poses of each of the bodies, a joint centering cost cannot be calculated directly through an inverse map, as it would make the problem nonconvex. Instead, we compare different proxies for a joint-centering cost.
One existing method is a posture cost, where we penalizing the deviation of each link to a desired posture, penalizing $\monogram{X}{W}{i}$ against $\monogram{X}{W}{i}_d$.
The second method is a more direct link cost, penalizing the desired posture of each body relative to its parent body, penalizing $\monogram{X}{i-1}{i}$ against $\monogram{X}{i-1}{i}_d$ for each body.
For the second method, we compare taking the $\ell_1$, $\ell_2$, and squared $\ell_2$ norm as a penalty.

\begin{table}[h]
    \renewcommand{\arraystretch}{1.25}
	\centering
	\begin{tabular}{|c|c|c|c|} \hline
        Cost & Runtime (sec) & True Cost & Success Rate\\ \hline
e        Posture Cost & 123 & 8.61 & 0.84\\ \hline
        Link Cost ($\ell_1$) & 131 & 6.06 & \multirow{3}{*}{0.88}\\ \cline{1-3}
        Link Cost ($\ell_2$) & 227 & 6.12 & \\ \cline{1-3}
        Link Cost ($\ell_2^2$) & 289 & 5.53 & \\ \hline
	\end{tabular}
	\caption{ 
        Average Performance of Global-IK with 70\% scene polytope coverage over 4 different proxy costs.
        The costs which more accurately approximate the desired quadratic joint-centering cost yield a higher runtime.
	}         
	\label{tab:global_ik_costs}
\end{table}

In \Cref{tab:global_ik_costs} we solve the Global-IK mixed-integer program to optimality for each cost over 100 target poses and compare with the true quadratic joint-centering cost that it achieves.
We also compare the success rate and runtime associated to each cost.
In summary, by combining Global-IK with a Clique Cover Method to generate collision-free polytopes in space using IRIS, we create a automated pipeline for global convex IK, and by using an $\ell_2^2$ norm link-cost, and we can accurately minimize a quadratic joint-centering cost equivalent to our nonconvex optimization approach.

\section{Further Details on Setup of Experiments}
\label{appx:experiment_implementation}
In this appendix, we give further details on the experiments performed in \Cref{sec:experiments}, including precise formulations of the optimization problems and descriptions of how targets are sampled.

\subsection{Common Costs and Constraints}
\label{appx:experiment_implementation:common_costs_and_constraints}

We frequently apply a quadratic joint centering cost of the form $q\tpose Mq$, where $M\in\R^{d\times d}$ is positive semidefinite.
We generally only penalize joints which have limits, as the goal of a joint centering cost is usually to avoid these limits.
We also consider costs like $(q-q_{\mathrm{nom}})\tpose M(q-q_{\mathrm{nom}})$, where $q_{\mathrm{nom}}$ is some nominal configuration.

Collision avoidance is perhaps the most intricate constraint.
We can efficiently compute the distance between two convex geometries, as well as the partial derivatives of that distance with respect to the position and orientation of those geometries.
We encode collision avoidance as the minimum distance between geometries being above some threshold, using Drake's \href{https://github.com/RobotLocomotion/drake/blob/v1.44.0/multibody/inverse_kinematics/minimum_distance_lower_bound_constraint.h}{\texttt{MinimumDistanceLowerBoundConstraint}}.
The individual pairwise distances are computed via FCL~\cite{pan2012fcl} and libccd~\cite{fiser2018libccd}.
For notational brevity, we let $\SDF(q)$ be the minimum pairwise signed distance, and write the constraint as $\SDF(q)\ge d_{\min}$.
Finally, \cite{beeson2015trac} shows that the choice of representation for the end-effector pose constraint in $\SE(3)$ (e.g. dual quaternions vs $4\times 4$ homogeneous matrices) affects the performance.
To facilitate direct comparisons between the formulations, we implement the constraint $\monogram{o}{W}{G} = \monogram{o}{W}{G}_{des}$ using the Euler angles representation for both formulations.

\subsection{Arm on a Table}
\label{appx:experiment_implementation:arm_on_a_table}

The optimization problem we solve for \Cref{sec:experiments:arm_on_a_table} is
\begin{subequations}
\label{eq:arm_on_a_table}
\begin{align}
        \min_q \colsep & q\tpose I_dq
            \label{eq:arm_on_a_table:cost}\\
        \ST \colsep & q\in\R^d
            \label{eq:arm_on_a_table:variables}\\
        \colsep & \monogram{X}{W}{G}(q)=\monogram{X}{W}{G}_{\des}
            \label{eq:arm_on_a_table:ik_constraint}\\
        \colsep & q_{\lb}\le q\le q_{\ub},
            \label{eq:arm_on_a_table:joint_limits}\\
        \colsep & \SDF(q)\ge d_{\min},
            \label{eq:arm_on_a_table:min_distnace}
\end{align}
\end{subequations}
where $I_d$ is the $d\times d$ identity matrix, $d_{\min}$ is set to $1$mm. 

The random end-effector targets $\monogram{X}{W}{G}_{\des}$ are chosen by uniformly sampling in joint-space within joint limits, throwing out the configurations that are not collision-free. This way, we guarantee that every optimization problem has a feasible solution.
Additionally, we pick initial guesses by sampling in joint space, except now allowing for joint configurations in collision and outside joint limits.
We match these sampled initial guesses between the old and new formulation by taking each sampled joint configuration $q_i$, and setting initial guesses for $\monogram{X}{W}{G}_i$, $\psi_i$, and $\kappa_i$ such that $q_i = \mathrm{IK}(\monogram{X}{W}{G}_i, \psi_i, \kappa_i)$ in the new formulation.
We sample 100 targets, and initial guesses, resulting in $10^4$ total trials.

For the sampling baseline, we fix the end-effector position in the analytic IK mapping and take samples linearly from $\psi\in[0,2\pi]$, $\kappa\in \ms K$ 
For the iiwa, $\card{\ms K}=8$.
We draw 500 samples, and choose the one which achieves the lowest cost while satisfying the constraints.
(This number was chosen to have a similar runtime to the corresponding optimization formulations, towards providing a fair comparison.)

\subsection{Grasp Selection}
\label{appx:experiment_implementation:grasp_selection}

Given a mug, with body frame $M$, we solve
\begin{subequations}
\label{eq:grasp_selection}
\begin{align}
        \min_q \colsep & q\tpose I_dq
            \label{eq:grasp_selection:cost}\\
        \ST \colsep & q\in\R^d
            \label{eq:grasp_selection:variables}\\
        \colsep & p_{\lb} \le \monogram{p}{M}{G}_M \le p_{\ub} \label{eq:grasp_selection:mug_constraint}\\
        \colsep & q_{\lb}\le q\le q_{\ub},
            \label{eq:grasp_selection:joint_limits}\\
        \colsep & \SDF(q)\ge d_{\min},
            \label{eq:grasp_selection:min_distance}
\end{align}
\end{subequations}
where $\monogram{p}{M}{G}_M = \monogram{X}{M}{W} \: \monogram{p}{W}{G}$ is the position of the center of the gripper relative to the mug, expressed in the frame of the mug.
To center the gripper along the central axis of the mug, we allow $p_{\lb}, p_{\ub} = [0, 0, \pm h]$, where $h = 0.035$ is a little less than half the height of the mug.
Since the first two components of $p_{\lb}$ and $p_{\ub}$ are equal, we have a nonlinear equality constraint in the old formulation.
However, in the new formulation, $\monogram{p}{M}{G}_M$ is a linear function of the end-effector pose $\monogram{X}{W}{G}$, so we have a relative interior.

To choose reasonable mug poses for the arm to target, we randomly choose a joint configuration for the iiwa, and place a mug centered at the finray gripper.
The orientation of the mug is aligned with the gripper so that the handle of the mug faces directly away from the iiwa. 
Although the collision geometry of the mug heavily restricts the feasible set, this ensures that the optimization problem is feasible.
Initial guesses are chosen randomly in configuration space, within the joint limits of the iiwa, not requiring them to be collision-free.
We choose 40 targets, and 40 initial guesses for 1600 trials. 

For the sampling IK baseline, we have five degrees of freedom, along with the discrete parameter $\kappa$.
This consists of three degrees of freedom for end-effector orientation, one for the position along the axis, and one for the self-motion parameter $\psi$.
As in \Cref{appx:experiment_implementation:arm_on_a_table}, we discretize each parameter into a grid, drawing samples linearly from $\psi \in [0, 2\pi]$, $\kappa \in \ms K$, as well as 
sampling $\monogram{o}{W}{G}$ by setting roll and yaw within $[-\pi, \pi]$, and pitch within $[-\pi/2, \pi/2]$.

\subsection{A Basic Mobile Manipulator}
\label{appx:experiment_implementation:kmr_iiwa}

We solve
\begin{subequations}
\label{eq:kmr}
\begin{align}
        \min_q \colsep & (q - q_{\mathrm{nom}})\tpose M_d(q - q_{\mathrm{nom}})
            \label{eq:kmr:cost}\\
        \ST \colsep & q\in\R^d
            \label{eq:kmr:variables}\\
        \colsep & \monogram{X}{W}{G}(q)=\monogram{X}{W}{G}_{\des}
            \label{eq:kmr:ik_constraint}\\
        \colsep & q_{\lb}\le q\le q_{\ub},
            \label{eq:kmr:joint_limits}\\
        \colsep & \SDF(q)\ge d_{\min}.
            \label{eq:kmr:min_distnace}
\end{align}
\end{subequations}
The primary difference between this problem and \eqref{eq:arm_on_a_table} is the cost function, where $M_d$ is the identity matrix, except the diagonal entries corresponding to the pose of the base are set to zero.
In this way, we only set a quadratic cost on the joint angles of the iiwa arm.

We choose to implement the joint centering cost relative to a nominal joint configuration.
This is because with a mobile base, imposing the cost $q\tpose I_dq$ would easily enter the robot into singularity, with the arm of the iiwa extending straight out to minimize the cost.
This is a larger issue for this experiment than previous ones, due to the greater degree of kinematic redundancy -- the arm can be near singularity and still achieve the desired end-effector pose by carefully positioning the remaining joints.

Since the obstacles in this setup take up a much higher fraction of configuration space, we sample collision-free $q$ for both initial guesses and targets.
This leads to much higher success rates, which we believe is due to configurations where the robot is deeply colliding with the table providing unclear gradient information.
To add to the collision-avoidance challenge, we sample targets close to the table.

For a given gripper target, there are four degrees of self-motion, which we parameterize as the pose of the base $(x,y, \theta)$ and the arm self-motion $\psi$, along with $\kappa$ as an additional discrete parameter.
We find that while sampling in this space does have a high success rate, it can no longer reliably find the global minimum.

\subsection{A Bimanual Mobile Manipulator}
\label{appx:experiment_implementation:pr2}

For the bimanual mobile manipulator, we let $\monogram{X}{W}{L}(q)$ and $\monogram{X}{W}{R}(q)$ denote the pose of the left and right hand, respectively, and $\monogram{X}{W}{L}_{\des}$ and $\monogram{X}{W}{R}_{\des}$ are the corresponding target poses.
We solve
\begin{subequations}
\label{eq:pr2}
\begin{align}
        \min_q \colsep & (q - q_{\mathrm{nom}})\tpose I_d(q - q_{\mathrm{nom}}) 
            \label{eq:pr2:cost}\\
        \ST \colsep & q\in\R^d
            \label{eq:pr2:variables}\\
        \colsep & \monogram{X}{W}{L}(q)=\monogram{X}{W}{L}_{\des}
            \label{eq:pr2:ik_constraint_left}\\
        \colsep & \monogram{X}{W}{R}(q)=\monogram{X}{W}{R}_{\des}
            \label{eq:pr2:ik_constraint_right}\\
        \colsep & q_{\lb}\le q\le q_{\ub},
            \label{eq:pr2:joint_limits}\\
        \colsep & \SDF(q)\ge d_{\min}.
            \label{eq:pr2:min_distnace}
\end{align}
\end{subequations}
Like in \Cref{sec:experiments:kmr_iiwa}, we impose a cost $(q - q_{\mathrm{nom}})\tpose I_d(q - q_{\mathrm{nom}})$ such that $M_d$ is the identity for the joint angles of the two arms and zero elsewhere.

One key difference between the arms of the PR2 and the KUKA iiwa we considered before is that the joint limits are much more restrictive.
While the iiwa's joint limits encompass around $25\%$ of $\mathbb T^7$, the joint limits for a single arm of the PR2 cover less than $2\%$.
This means that within the subset of configuration space for the two arms $\mathbb T^{14}$, less than $0.05\%$ is valid without even considering self-collisions. 
Additionally, instead of the shoulder-elbow-wrist redundancy parametrization of the iiwa, which is known to be favorable~\cite{elias2024stereographicsew}, the PR2's analytic IK function uses the first joint angle as the redundancy parameter.
This joint specifically has restrictive joint limits, so the optimizer can easily get stuck in a local minimum outside of these limits.
We label the redundancy parameters of the left and right arm $\psi_r$ and $\psi_l$ respectively. 

We sample collision-free configurations to get both 40 targets and 40 initial guesses, as in \Cref{sec:experiments:kmr_iiwa}, with a total of 1600 trials. 
In order to get any feasible solutions via sampling, we restrict the base to be near the midpoint of the left and right target, but still sample in six dimensions of self-motion ($x,y,\theta$ for the base, $z$ for the torso lift joint, and $\psi_r, \psi_l$ for the arms themselves). Redundancy parameters $\psi_r$ and $\psi_l$ are taken uniformly between the lower and upper joint limits of the shoulder joint. $x$ and $y$ are chosen within a range of $\pm 0.5$ around the midpoint of the two targets.

\subsection{Humanoid Stability}
\label{appx:experiment_implementation:hubo}

The IK problem considered in this experiment is
\begin{subequations}
\label{eq:hubo}
\begin{align}
    \find \colsep & q\in\R^{d+1},
        \label{eq:hubo:variables}\\
    \ST \colsep & \monogram{X}{W}{R}_{\operatorname{hand}}(q)=\monogram{X}{W}{R}_{\operatorname{hand},\des},
        \label{eq:hubo:ik_constraint}\\
    \colsep & [p(\monogram{X}{W}{L}_{\operatorname{foot}}(q))]_{\operatorname{z}}=0,
        \;[p(\monogram{X}{W}{R}_{\operatorname{foot}}(q))]_{\operatorname{z}}=0,
        \label{eq:hubo:foot_position}\\
    \colsep & [o(\monogram{X}{W}{L}_{\operatorname{foot}}(q))]_{\operatorname{rp}}=0,
        \;[o(\monogram{X}{W}{R}_{\operatorname{foot}}(q))]_{\operatorname{rp}}=0,
        \label{eq:hubo:foot_orientation}\\
    \colsep & q_{\lb}\le q_{0:7}\le q_{\ub},
        \label{eq:hubo:joint_limits}\\
    \colsep & \norm{q_{0:4}}_2^2=1,
        \label{eq:hubo:unit_quaternion}\\
    \colsep & \SDF(q)\ge d_{\min},
        \label{eq:hubo:min_distance}\\
    \colsep & [p_{\com}(q)]_{\operatorname{xy}}\in\mc P(q)=\Conv\set{[p_{\operatorname{foot},i}(q)]_{\operatorname{xy}}}_{i=1}^8.
        \label{eq:hubo:stability}
\end{align}
\end{subequations}
We now further discuss the notation used, along with several other aspects of this optimization problem.
We use $[\cdot]_{\operatorname{xy}}$ to denote just the $\operatorname{x}$ and $\operatorname{y}$ components of a position in 3D space (in the world frame), and similarly $[\cdot]_{\operatorname{z}}$ denotes the $\operatorname{z}$ component.
$[\cdot]_{\operatorname{rp}}$ denotes just the roll and pitch components of a set of Euler angles.

The configuration vector $q$ encodes both the 6DoF pose of the torso, as well as all the remaining joints discussed before.
The orientation of the torso is represented by a unit quaternion $q_{0:4}$, with the added unit-norm constraint \eqref{eq:hubo:unit_quaternion}.
(For the new formulation, we replace the unit quaternion with a set of Euler angles in order to remove the nonlinear equality constraint.)
No restrictions are placed on the torso position, but its horizontal position is indirectly constrained by the right-hand target pose \eqref{eq:hubo:ik_constraint}, and its vertical position is indirectly constrained by the requirement that the feet are flush with the ground \eqref{eq:hubo:foot_position}, \eqref{eq:hubo:foot_orientation}.

The position of the center of mass of the robot is computed a function of the configuration vector $q$.
The pose of each link is determined as a function of $q$ with forward kinematics, and then $p_{\com}(q)$ is the weighted sum of the centers of mass of the links, which is implemented in Drake as the function \href{https://github.com/RobotLocomotion/drake/blob/v1.44.0/multibody/plant/multibody_plant.h#L3776}{\texttt{CalcCenterOfMassPositionInWorld}}.
The support polytope $\mc P(q)$ is the convex hull of the eight foot points $p_{\operatorname{foot},i}(q)$ (four per foot, one at each corner).%

Stability is actually enforced by \eqref{eq:hubo:stability}.
One way to write this constraint is to introduce convex hull multipliers $\lambda_{i}\in[0,1]$, $i\in[8]$, and add the constraints
\begin{subequations}
\label{eq:stability_equality}
\begin{align}
    & \sum_{i=1}^8\lambda_i=1,
        \label{eq:stability_equality:multipliers}\\
    & \sum_{i=1}^8\lambda_i [p_{\operatorname{foot},i}(q)]_{\operatorname{xy}}=[p_{\com}(q)]_{\operatorname{xy}}.
        \label{eq:stability_equality:convex_hull}
\end{align}
\end{subequations}
\eqref{eq:stability_equality:multipliers} is a linear equality constraint, but \eqref{eq:stability_equality:convex_hull} is a nonlinear equality constraint.
This formulation allows us to study the performance of the new formulation in the presence of additional nonlinear equality constraints.

However, the feasible set is still positive measure in the new formulation, and in fact, it is possible to rewrite the constraint using only inequality constraints.
According to Carath\'eodory's theorem, a point $p\in\R^d$ is in the convex hull of a set of points $v_i$ if and only if it is contained in some $(d+1)$-simplex with vertices $v_{i_1},\ldots,v_{i_{d+1}}$~\cite{caratheodory1911convexhullsimplex}.
Normally, such a result is not computationally useful, as there are $\binom{N}{d+1}$ possible simplices, but for our fixed problem data, there are only $\binom{8}{3}=56$ triangles in 2D space.
Thus, we can enforce that the center of mass lie in at least one of these triangles, avoiding any nonlinear equality constraints.

We can compute the signed distance from a point $p$ to a triangle $\triangle v_1v_2v_3$ by first computing vectors $n_{12}^\perp=v_2-v_1$, $n_{23}^\perp=v_3-v_2$, and $n_{31}^\perp=v_1-v_3$, then computing normals $n_{12}=(-[n_{12}^\perp]_{\operatorname{y}}, [n_{12}^\perp]_{\operatorname{x}})$ (and similarly for $n_{23}$ and $n_{31}$).
Define the ``slacks''
\begin{equation*}
    s_{12}:=\frac{n_{12}\tpose(p-v_1)}{\norm{n_{12}}},
    s_{23}:=\frac{n_{23}\tpose(p-v_2)}{\norm{n_{23}}},
    s_{31}:=\frac{n_{31}\tpose(p-v_3)}{\norm{n_{31}}}.
\end{equation*}
Now, consider
\begin{equation}
    \label{eq:ordered_triangle_slack}
    s(p,v_1,v_2,v_3):=\min\set{s_{12},s_{23},s_{31}}.
\end{equation}
If $p\not\in\triangle v_1v_2v_3$, then at least one of $s_{ij}$ will be positive and at least one will be negative, and thus, $s$ will be negative.
If $p\in\triangle v_1v_2v_3$ and the points are given in counterclockwise order, then the normals $n_{ij}^\perp$ will be outward pointing, so each $s_{ij}\ge 0$, so $s\ge 0$.
But if the points are in clockwise order, the normals are inward pointing, so each $s_{ij}\le 0$, so $s\le 0$.
Thus, $s\ge 0$ is sufficient, but not necessary, for $p\in\triangle v_1v_2v_3$.

We can fix this issue by taking checking $\triangle v_1v_3v_2$, which has the opposite winding.
If we let $s_{ij}'$ be the slacks for $\triangle v_1v_3v_2$, it can be shown that $(s_{12}',s_{23}',s_{31}')=(-s_{12},-s_{23},-s_{31})$.
This means if $p\in\triangle v_1v_2v_3$, then one of $s\ge 0$ or $s'\ge 0$, and if $p\not\in\triangle v_1v_2v_3$, then $s<0$ and $s'<0$.
Thus,
\begin{equation}
    \label{eq:triangle_containment_condition}
    \max\set{s(p,v_1,v_2,v_3),s(p,v_1,v_3,v_2)}\ge 0
    \;\Leftrightarrow\;
    p\in\triangle v_1v_2v_3
\end{equation}
and we write our stability constraint in inequality form as
\begin{equation}
    \label{eq:stability_inequality}
    \max_{1\le i<j<k\le 8}\set{s(p,v_i,v_j,v_k),s(p,v_i,v_k,v_j)}\ge 0,
\end{equation}
where $v_1,\ldots,v_8$ are the eight foot points.

To determine the goal end-effector targets for our experiments, we begin with a nominal stable configuration, shown in \Cref{fig:experimental_setup:hubo}.
We then move the torso to a uniformly random position within $\pm 1$ meter in the $\operatorname{x}$ and $\operatorname{y}$ directions, and between $0.7$ and $0.9$ meters in the $\operatorname{z}$ direction, and the yaw is uniformly sampled from the range $[-\pi,\pi)$.
Finally, the right arm has its joint angles uniformly sampled from within its limits.
This process is repeated until the resulting configuration is collision-free, and then the pose of the right hand is used as the target.
This ensures that it is possible for the humanoid to reach the goal configuration.

For this experiment, we found it necessary to enforce the reachability constraint at every solver iteration.
Current optimization software does not generally allow the user to specify such additional information.
Thus, we utilize two strategies to more aggressively enforce these constraints.

The first strategy is to provide an initial guess to the optimizer which is feasible for at least the hard constraints.
In particular, the target pose must be reachable from the base of the kinematic chain and (along with the self-motion parameters) not yield a singular joint configuration.
In general, this is not difficult, since one can just sample from the interior of the joint limits and run forward kinematics to get the initial value.
We do not have to worry about sampling a singular configuration, since they are measure zero.

The second trick is to add a cost of $-\mu\log(f(x))$ to our formulation for any hard constraints $f(x)\ge 0$ for a small, user-defined parameter $\mu$.
The nature of $\log(x)$ as having $x>0$ as its domain forces the line-search in solvers such as IPOPT to remain in the interior of the feasible domain.
Moreover, provided that $f$ is sufficiently smooth, the gradient and Hessian of $\log(f(x))$ will vary smoothly as well.
This cost effectively acts a soft, differentiable indicator function on the set of hard-feasible constraints and is a standard method for enforcing constraints in nonlinear optimization~\cite{wachter2006ipopt}.
For the humanoid stability experiment, we add such costs to the reachability constraints and to avoid certain singular configurations of the arms which frequently appeared in our experiments.

\subsection{2D Scaling}
\label{appx:experiment_implementation:scaling_2d}

Our inverse kinematics condition requires the last link ends up at $(x_{\des}, y_{\des},\theta_{\des}) = \monogram{X}{W}{n}_{\des}$. 
We solve the following optimization problem, as well as the equivalent feasibility problem without the cost:
\begin{subequations}
\label{eq:2d}
\begin{align}
        \min_q \colsep & (q - q_{\mathrm{nom}})\tpose I_n(q - q_{\mathrm{nom}})
            \label{eq:2d:cost}\\
        \ST \colsep & q\in\R^n
            \label{eq:2d:variables}\\
        \colsep & \monogram{X}{W}{n}(q)=\monogram{X}{W}{n}_{\des}
        \label{eq:2d:pos_constraint}
\end{align}
\end{subequations}

The new formulation uses $q_0 \ldots q_{n-4}$ and $\monogram{X}{W}{n} = (\monogram{p}{W}{n}, \theta)$ as decision variables.
Solving backwards from the last link, our IK function computes $\monogram{p}{W}{n-1}_d$.
Then, the joint $q_{n-2}$ can be calculated as
\begin{equation}\label{eq:2d:ik}
    q_{n-2} = \pm 2\cos^{-1}(\Vert\monogram{p}{W}{n-1}_d - \monogram{p}{W}{n-3}\Vert \cdot n/2)
\end{equation}
The remaining joint angles $q_{n-3}, q_{n-1}$ can also be calculated similarly.
Thus, given $\monogram{X}{W}{n-3}$, reachability is equivalent to requiring \begin{equation}
    \mathcal D(q)= \Vert\monogram{p}{W}{n-1}_d - \monogram{p}{W}{n-3}\Vert_2^2 \cdot n^2/4 \le 1.
\end{equation}

We sample 300 targets per number of links uniformly within $\Vert\monogram{p}{W}{n}_{\des}\Vert < 1 - 2/n$, which guarantees a valid target.
Initial guesses and $q_{\mathrm{nom}}$ are chosen independently at random so that the optimizer does not start at a global minimum.

We found that for IPOPT, setting a bounding box for the joints $q_i$ was necessary to find a reasonable cost.
However, implementing the joint limits $-\pi \le q_i \le \pi$ resulted in low success rates, as the optimizer often ended up at the wrong end of the joint limits and got stuck at an infeasible local minimum.
We opted to implement the bounding box of $-2\pi \le q_i \le 2\pi$ for both formulations.
Both formulations were able to achieve more than $95\%$ success rates with this addition.

\subsection{3D Scaling}
\label{appx:experiment_implementation:scaling_3d}

We characterize our $n+7$ link arm with the DH parameters listed in \Cref{tab:DH_parameters}.
\begin{table}[!h]
    \centering
    \begin{tabular}{|c|c|c|c|}\hline
     link & $d_i$ & $\alpha_i$ & $a_i$ \\ \hline
     0 & $l$ & $-\pi/2$ & 0 \\ \hline
     1 & 0 & $\pi/2$ & 0 \\ \hline
     2 & $l$ & $-\pi/2$ & 0 \\ \hline
     3 & 0 & $\pi/2$ & 0 \\ \hline
     4 & $l$ & $-\pi/2$ & 0 \\ \hline
     5 & 0 & $\pi/2$ & 0 \\  \hdashline
     \multicolumn{4}{|c|}{$\vdots$} \\  \hdashline
     $n$ & $l$ & $-\pi/2$ & 0 \\ \hline
     $n + 1$ & 0 & $\pi/2$ & 0 \\\hline
     $n+ 2$ & $l$ & $\pi/2$ & 0 \\ \hline
     $n + 3$ & 0 & $-\pi/2$ & 0 \\\hline
     $n + 4$ & $l$ & $-\pi/2$ & 0 \\ \hline
     $n + 5$ & 0 & $\pi/2$ & 0 \\\hline
     $n + 6$ & $l$ & $0$ & 0 \\\hline
\end{tabular}
    \caption{DH parameters for scaled 3d arm.}
    \label{tab:DH_parameters}
\end{table}

$l$ is chosen so that the total length of the arm is constant. 
The first $n$ links alternate between two types of links, followed by the DH parameters of the KUKA iiwa (except with different link lengths).
We can thus use the analytic IK function of the iiwa given $\monogram{X}{W}{n}$ to determine reachability and the last seven joint angles.

We solve the optimization problem
\begin{subequations}
\label{eq:3d}
\begin{align}
        \min_q \colsep & q\tpose I_dq
            \label{eq:3d:cost}\\
        \ST \colsep & q\in\R^d,
            \label{eq:3d:variables}\\
        \colsep & \monogram{X}{W}{n + 6}(q)=\monogram{X}{W}{n + 6}_d,
            \label{eq:3d:pos_constraint} \\ 
        \colsep & -\pi \le q_i \le \pi. \label{eq:3d:joint_limits}
\end{align}
\end{subequations}
For the new formulation, we replace decision variables $\{q_{n}, \ldots, q_{n+6}\}$ with $IK(\monogram{X}{n}{n+6}, \psi, \kappa_0)$, choosing $\kappa_0$ at random. 
We create 200 feasible targets per number of links, by sampling Euclidean poses randomly, choosing $-0.3 \le x,y,z \le 0.3$ and a random orientation.
Initial guesses are joint angles chosen independently at random.

\end{document}